\documentclass[lettersize,journal]{IEEEtran}
\usepackage{amsmath,amsfonts}
\usepackage{algorithmic}
\usepackage{algorithm}
\usepackage{array}
\usepackage[caption=false,font=normalsize,labelfont=sf,textfont=sf]{subfig}
\usepackage{textcomp}
\usepackage{stfloats}
\usepackage{url}
\usepackage{verbatim}
\usepackage{graphicx}
\usepackage{cite}

\usepackage{booktabs}
\usepackage{multirow}
\usepackage{amssymb}

\usepackage[dvipsnames]{xcolor}
\definecolor{R1}{RGB}{220,34,48}
\definecolor{R2}{RGB}{34,48,220}
\definecolor{R3}{RGB}{51,158,0}
\begin{document}

\title{Face De-identification: State-of-the-art Methods and Comparative Studies}

\author{Jingyi Cao, Xiangyi Chen, Bo Liu,~\IEEEmembership{Senior Member,~IEEE,} Ming Ding,~\IEEEmembership{Senior Member,~IEEE,} Rong Xie,~\IEEEmembership{Member,~IEEE,} Li Song,~\IEEEmembership{Senior Member,~IEEE}, Zhu Li,~\IEEEmembership{Senior Member,~IEEE,} Wenjun Zhang,~\IEEEmembership{Fellow,~IEEE}
\thanks{Jingyi Cao, Xiangyi Chen, Rong Xie, Li Song and Wenjun Zhang are with The Institute of Image Communication and Network Engineering, Shanghai Jiao Tong University, Shanghai 200240, China (email: cjycaojingyi@sjtu.edu.cn;chenxiangyi@sjtu.edu.cn;
xierong@sjtu.edu.cn; song li@sjtu.edu.cn; zhangwenjun@sjtu.edu.cn). }
\thanks{Bo Liu is with Australian Artificial Intelligence Institute and School of Computer Science, University of Technology
Sydney, Ultimo NSW 2007, Australia (email:bo.liu@uts.edu.au).}
\thanks{Ming Ding is with Data61, CSIRO, Sydney NSW, Australia (ming.ding@data61.csiro.au).}
\thanks{Zhu Li is with the Department of Computer Science and Electrical Engineering, University of Missouri-Kansas City, Kansas City, MO 64110 USA
(e-mail: lizhu@umkc.edu).}
}







\markboth{Journal of \LaTeX\ Class Files,~Vol.~14, No.~8, August~2021}%
{Shell \MakeLowercase{\textit{et al.}}: A Sample Article Using IEEEtran.cls for IEEE Journals}


\maketitle

\begin{abstract}

The widespread use of image acquisition technologies, along with advances in facial recognition, has raised serious privacy concerns. Face de-identification usually refers to the process of concealing or replacing personal identifiers, 
which is regarded as an effective means to protect the privacy of facial images. 
A significant number of methods for face de-identification have been proposed in recent years. In this survey, we provide a comprehensive review of state-of-the-art face de-identification methods, categorized into three levels: pixel-level, representation-level, and semantic-level techniques. We systematically evaluate these methods based on two key criteria, the effectiveness of privacy protection and preservation of image utility, highlighting their advantages and limitations. Our analysis includes qualitative and quantitative comparisons of the main algorithms, demonstrating that deep learning-based approaches, particularly those using Generative Adversarial Networks (GANs) and diffusion models, have achieved significant advancements in balancing privacy and utility. Experimental results reveal that while recent methods demonstrate strong privacy protection, trade-offs remain in visual fidelity and computational complexity. This survey not only summarizes the current landscape but also identifies key challenges and future research directions in face de-identification.
\end{abstract}


\section{Introduction}
\label{sec:introdection}
The rapid proliferation of artificial intelligence (AI) technologies, particularly deep learning-based models, has significantly transformed the landscape of face recognition systems. These advancements have brought about a wide range of applications, such as social media platforms. Unfortunately, 
accompanying these flourishing advancements, 
privacy concerns have also grown fast. 
Especially, 
faces are regarded as one of the most privacy-sensitive biological information directly related to personal identity. 
The essence of face recognition is biometric authentication, 
whose characteristics are unique and irrevocable. 
Once face recognition technology is used for cross-referencing with other databases, 
it will further disclose the user's other sensitive information. 
The previous study \cite{acquisti2014face} has shown how faces can be used as the link across different databases and the trails associated with their different personas, thus violating privacy.

Facial privacy issues are receiving more attention these days. 
Restrictive laws and regulations such as the General Data Protection Regulations (GDPR) \cite{houser2018gdpr} have taken effect, 
which stipulates that data collection, 
sharing, 
and analysis by companies without the user’s knowledge is considered illegal and regular consent is required for any use of their personal data to ensure data privacy. 
In GDPR, 
privacy information was defined as ``personal data that are related to an identified or identifiable natural person", 
so personal identity is the most important part of facial image protection. 
To address this privacy issue, face de-identification (aka. face anonymization and face obfuscation) techniques have emerged as a crucial means of protecting privacy by anonymizing facial features in images without compromising their utility for non-identification tasks.

Face de-identification refers to the process of concealing or altering personally identifiable facial features to prevent recognition. These techniques have a broad range of applications, from protecting individuals' identities in media interviews and video surveillance \cite{senior2009privacy}, and medical research~\cite{zhu2020deepfakes}, to ensuring privacy in surveillance footage and social platforms~\cite{lin2021fpgan, lee2021development}. To date, many face de-identification methods have been proposed, which aim at protecting facial sensitive information (especially identity) while preserving utility for identity-unrelated applications. In addition, several related surveys have been conducted on related topics over the past decade, 
which are outlined in Table~\ref{tab:related-survey}.
\begin{enumerate}
    \item Padilla et al. \cite{padilla2015visual} provided a review and classification of visual privacy protection methods, as well as an analysis of how visual privacy protection techniques are deployed in existing privacy-aware intelligent monitoring systems.
    
    \item Ribaric et al. \cite{ribaric2015overview} reviewed existing face de-identification in still images and videos at the time, namely non-deep-learning image filtering methods and $k$-Same based methods.
    
    \item Ribaric et al. \cite{ribaric2016identification} presented a review of de-identification methods for non-biometric identifiers, physiological, behavioral and soft-biometric identifiers in multimedia contents.
    
    \item Liu et al. \cite{liu2020privacy} targeted privacy issues in dynamic Online Social Networks from a user-centric perspective. They proposed a privacy analysis framework consisting of three stages with different principles, observable privacy, contextual privacy and inferential privacy.
    
    \item Cai et al. \cite{cai2021generative} systematically summarized the application of GAN in privacy and security, which discussed privacy from the perspectives of both data type and model and security from model robustness, malware detection, etc. They also provided unsolved challenges in the application scenario, model design and data utilization.
    
    \item The relation between privacy and machine learning has been discussed in \cite{liu2021machine}, which covers three categories: private machine learning, machine learning-aided privacy protection and machine learning-based privacy attack and corresponding methods.
    
    \item The proposed taxonomy of \cite{meden2021privacy} was tied to biometric recognition systems and partitioned biometric privacy-enhancing techniques into image-level, representation-level and inference-level, which also reviewed existing datasets, relevant standards and regulations. 
    
    \item Shopon et al. \cite{shopon2021biometric}  summarized fragmented research on biometric de-identification and provided a classification mechanism based on the modalities employed and the types of biometric traits preserved after de-identification. They also discussed their applications in various domains such as cybersecurity.
    
    \item The survey \cite{wenger2021sok} started with a face recognition system and classified anti-facial recognition tools according to their targets, from data collection and model training to inference. They created a systematic framework to analyze the benefits and trade-offs of different AFR approaches and then considered the technical and societal challenges.
    
    \item Zhang et al. \cite{zhang2022visual} reviewed privacy attack methods and corresponding defense mechanisms in both visual data and visual systems under the context of deep learning.
    
    \item Park et al. \cite{park2022current} focused on reviewing GAN-based facial de-identification algorithms. Particularly, their study evaluated existing methods in terms of data utility pertaining to the preservation of dermatologically interesting features such as skin color, pigmentation and texture.

    \item Maity et al. \cite{maity2023preserving} conducted a thorough exploration into the intersection of video analytics and privacy preservation, focusing on two core techniques: face de-identification and background blurring. They rigorously analyze the latest advancements, highlight both their efficacy and inherent limitations, so as to underscore the practical significance of these techniques through real-world applications in surveillance and the dynamic landscape of social platforms.
\end{enumerate}

\begin{table*}[tb]
\centering
  \caption{Summary on prior survey articles}
  \label{tab:related-survey}
  \resizebox{0.8\textwidth}{!}{
  \begin{tabular}{cc}
    \toprule
    Ref & Focuses \\
    \midrule
    \cite{padilla2015visual} & Visual Privacy Protection Methods and Privacy-Aware Monitoring Systems \\
    \cite{ribaric2015overview} & Traditional Face De-identification Methods \\
    \cite{ribaric2016identification} & Privacy Protection in Multimedia Contents\\
    \cite{liu2020privacy} & Image Privacy in Online Social Networks\\
    \cite{cai2021generative} & Generative Adversarial Networks in Privacy and Secure Applications\\
    \cite{liu2021machine} & Interactions between Privacy and Machine Learning\\

    \cite{meden2021privacy} & Face Biometric Privacy-Enhancing Techniques\\
    \cite{shopon2021biometric} & Biometric Systems De-Identification \\
    \cite{wenger2021sok} &  Benefits and Trade-offs of Different Anti-Facial Recognition Technology\\
    \cite{zhang2022visual} & Visual Privacy Attack and Defense Methods \\
    \cite{park2022current} & GAN-based De-Identification Methods in Dermatology Use Cases \\
        \cite{maity2023preserving} & Privacy preservation in video analytics \\
    This Survey & Face De-identification Methods  \\
  \bottomrule
\end{tabular}}
\end{table*}

Despite the development of numerous techniques, the field still lacks a comprehensive survey that categorizes these methods and systematically evaluates their performance under different criteria. This paper aims to fill that gap by presenting a state-of-the-art review of face de-identification methods. We categorize existing approaches based on their operational levels (pixel, representation, and semantic) and assess their strengths and limitations in terms of privacy protection and image utility preservation. 

Our key contributions are summarized as follows.
\begin{itemize}
    \item We present a state-of-the-art survey on face de-identification, 
    categorizing existing works by different image processing levels, 
    i.e., pixel-level, representation-level, and semantic-level, 
    and further discuss the characteristics of each category.
    \item We summarize relevant metrics from the perspectives of privacy protection and image utility, 
    and propose a comprehensive evaluation framework for de-identification algorithms.
    \item We perform qualitative and quantitative experimental comparisons of several algorithms to discuss their advantages and disadvantages. 
    \item We summarize the analysis and comparative research of face de-identification methods, 
    and further point out open problems and possible future research directions in this field.
\end{itemize}


The remainder of this paper is organized as follows. Section~\ref{sec:preliminary} reviews preliminary knowledge on face recognition and de-identification technologies. Subsequently, 
we review the existing face de-identification methods in Section~\ref{sec:methods} and classify them into \textit{pixel-level}, \textit{representation-level}, and \textit{semantic-level}, 
which can help readers gain a high-level understanding of current work and basic ideas. 
In Section~\ref{sec:applications}, 
we show practical applications of face de-identification, 
including usage in specific domains and identity-agnostic computer vision tasks.
Afterwards, the evaluation criteria commonly used in face de-identification algorithms are summarized in Section~\ref{sec:evaluation-metrics}. 
Additionally, 
quantitative and qualitative studies are implemented to compare representative methods in Section~\ref{sec:experiments} and the results provide intuitions and insights of of existing algorithms to readers. 
Finally, 
we propose some future directions in our view for this task in Section~\ref{sec:future-direction} and conclude our work in Section~\ref{sec:conclusion}.

\section{Preliminary}
\label{sec:preliminary}
This section introduces the preliminary knowledge in the research of face de-identification, reviews the research content of face recognition and de-identification, and introduces the principles of deep learning technology and privacy protection theory.
\subsection{Face Recognition and De-identification}
\label{sec:recognition-and-de-id}
\subsubsection{Face Recognition}
Face recognition is a biometric technology that automatically identifies individuals based on their facial features, including statistical and geometric characteristics. It is one of the most important applications of image analysis and understanding. Face recognition encompasses two main types of tasks: binary classification (face verification) and multi-classification (face retrieval). The term "face recognition" typically refers to identity recognition and verification based on optical facial images. The face recognition process can be summarized as a series of steps: first, a computer analyzes a facial image or video; next, it detects and tracks faces within the input; then, it extracts effective facial features; and finally, it determines the identity of the face by comparing these features against known identities.

Face recognition research, dating back to the late 1960s, initially focused on designing feature extractors coupled with machine learning algorithms for classification.  Traditional methods rely on hand-made features like edge texture description, and combine with machine learning techniques such as principal component analysis, linear discriminant analysis, and support vector machines. Early geometric-based approaches extracted contours and relationships of facial components, using these as feature vectors. While these methods offered fast recognition and low memory requirements, they often overlooked subtle local features, resulting in information loss. The latest feature point detection technology still struggles to meet accuracy requirements, highlighting the ongoing challenges in the field. 

Deep learning techniques have revolutionized face recognition, significantly improving accuracy and robustness by learning representations from large datasets that account for various conditions like lighting, posture, and expressions. For example, 
DeepFace \cite{DeepFace} introduced 3D model-based alignment, establishing a foundation for deep learning in this field. DeepID \cite{deepid, deepid2, DeepID3} series evolved from multi-classification to combining recognition and verification features, later incorporating advanced network architectures. FaceNet \cite{facenet} employed triplet loss to map faces to Euclidean space, where distances represent image similarity. Then, SphereFace \cite{sphereface} and CosFace \cite{cosface} introduced angular and cosine margins respectively, mapping features to hyperspherical space. Currently, ArcFace \cite{arcface} is considered the state-of-the-art, using additive angular margins for tighter feature distributions and clearer decision boundaries. Although current face-recognition models achieve impressive accuracy on specific datasets, they still face significant challenges. Primary among these are the influences of illumination and posture on recognition accuracy. Furthermore, cross-racial and cross-age recognition remains important areas for ongoing research and improvement.

\subsubsection{Face De-identification}

The concept of de-identification, a key aspect of privacy protection, lacks a consistent definition in existing literature. However, Ribaric et al. \cite{ribaric2016identification} provided a widely referenced definition, describing de-identification in multimedia content as \textit{``the process of concealing or removing personal identifiers, or replacing them with surrogate personal identifiers in multimedia content"}.

The main purpose of face de-identification is to conceal the identity information of a face image. For the given image $X$, the de-identification algorithm $\mathcal{F}$ aims to deceive the face recognition model $\mathcal{R}$ or human by reducing recognition accuracy, which can be expressed as,
\begin{equation}
    \mathcal{R}(X) \neq \mathcal{R}(\mathcal{F}(X)),
\end{equation}
where $\mathcal{R}(X)$ indicates the identity of $X$. 

In the past few years, face de-identification methods have evolvied from initial traditional approaches to more sophisticated deep learning-based techniques. Early methods focused on simple perturbation operations applied to facial regions. However, deep learning has enabled more advanced approaches, significantly enhancing the quality of de-identified results. The major challenge of face de-identification task lies in balancing privacy with utility. This trade-off is inherent to the task: as privacy protection effectiveness increases, the utility of de-identified results typically decreases.


The state-of-the-art deep learning-based methods 
can generate more realistic face de-identification results while ensuring the effectiveness of privacy protection, and they are the focus of the discussion in Section~\ref{sec:methods}.  

\subsection{Technical Background}
\label{sec:techniques}
In this section, we introduce the basic knowledge of deep learning technologies mainly used in face de-identification, especially Generative Adversarial Networks and Adversarial Perturbation. 


\subsubsection{Generative Adversarial Networks (GANs)} The generative adversarial Networks are based on adversarial training proposed by Goodfellow et al \cite{goodfellow2014generative}, which is used to generate images from random noise.
Tero Karras et al. \cite{karras2017progressive} was the first to apply GAN to face images and generated results with celebrity characteristics and satisfactory visual perception. 

\textbf{Basic Theory.} Generative Adversarial Networks are inspired by the game theory, which include two main parts: the generator $G$ and the discriminator $D$. The basic idea is to train through the confrontation between them to achieve the Nash equilibrium. The generator is trained to generate as realistic images as possible to deceive the discriminator, while the discriminator seeks to accurately distinguish the generated fake image from the real image. The corresponding optimization value function can be expressed as,

\begin{equation}
\begin{aligned}
\min _{G} \max _{D} V(G, D) = & \, \mathbb{E}_{x \sim P_{\text {data }}}\left[\log D(x)\right] \\
& + \mathbb{E}_{z \sim P_{z}}\left[\log \left(1-D(G(z))\right)\right].
\end{aligned}
\end{equation}

where $P_{\text{data}}$ represents the distribution of real data sets, $P_{G}$ is the distribution of network-generated images, and $z$ is the random noise in latent space. Under the optimal discriminator, the process of solving the optimal generator is essentially looking for the minimum distance between the distribution of real data and generated data.

\textbf{Conditional GANs.} The classical generative adversarial network generates results by randomly sampling from a priori distribution, which lacks controllability in the generation process. Conditional GANs (cGANs) introduce the class information $y$ into the training process, and the objective function can be formulated as,

\begin{equation}
\label{equ:cgan}
\begin{aligned}
\min _{G} \max _{D} V(G, D) = & \, \mathbb{E}_{x \sim P_{\text {data }}}\left[\log D(x \mid y)\right] \\
& + \mathbb{E}_{x \sim P_{G}}\left[\log \left(1-D(G(z \mid y))\right)\right].
\end{aligned}
\end{equation}

Conditional GANs \cite{mirza2014conditional} are widely used in image conversion tasks such as face attribute editing \cite{he2019attgan, choi2018stargan, liu2019stgan}. 
Some face de-identification algorithms (Section~\ref{sec:attributes}) treat facial attributes as privacy-sensitive information and achieve image protection through attribute editing. What's more, many face de-identification frameworks are designed based on conditional GANs or the idea of introducing conditional information, like \cite{gu2020password, deepprivacy, ciagan}.

\textbf{Progressive GANs.} Progressive GAN was proposed in Pro-GAN \cite{karras2017progressive}, which starts with a small input in low resolution, then adds new layer blocks, increases the output size of the generator and the input of the discriminator until the desired image size is reached. Progressive GAN is a training methodology of increasingly learning from low to high resolution, which can allow stable training with high-quality generated results. This strategy is also popular in network design and model training in GAN-based face de-identification methods.

\textbf{StyleGAN.} Inspired by style transfer, StyleGAN \cite{karras2019style} designed a new generator network structure, which can decouple the high-level semantic attributes of the image through unsupervised automatic learning. StyleGAN2 \cite{karras2020analyzing} analyzed the effectiveness of different resolution layers and designed a better network where the latent space can be more disentangled and the quality of the generated image can be higher. 
StyleGAN3 \cite{karras2021alias} fundamentally solved the problem of adhesion between coordinates and features and realized the invariance of image translation and rotation. Due to the powerful image generation and editing capabilities, many face swapping or face de-identification algorithms select StyleGAN as the generator \cite{karras2019style,nitzan2020face,jeong2021ficgan}.

\subsubsection{Adversarial perturbation}
Adversarial attack refers to adding adversarial noise to original samples to mislead the deep neural networks to output wrong prediction results, while humans cannot perceive the obfuscation. From different dimensions, adversarial examples can be categorized into false positive attacks and false negative attacks, white-box attacks and black-box attacks, target attacks and non-targeted attacks, and one-time attacks and iterative attacks \cite{yuan2019adversarial}. Szegedy et al. \cite{kurakin2018adversarial} first proposed L-BFGS method to generate adversarial examples against DNNs. Another classic algorithm is FGSM (Fast Gradient Sign Method) \cite{fgsm} proposed by GoodFellow et al., which is a gradient-based attack method based on constant model network parameters. The process is calculating the derivative of the model and adding noise in the gradient direction. Since the amplitude of perturbation is hard to train, I-FGSM (Iterative Fast Gradient Sign Method) \cite{i-fgsm} has been proposed to add noise by iterative methods. There are other adversarial perturbation methods including RAND-FGSM \cite{tramer2017ensemble}, JSMA \cite{papernot2016limitations}, DeepFool \cite{moosavi2016deepfool}, etc.

The targets of the existing privacy protection approach based on adversarial perturbation technology can be divided into two aspects: face detection and face recognition, where the former aims to prevent the face in the image from being caught and the latter is mainly to prevent the identification. 

\subsubsection{Diffusion Models}
Diffusion models have recently become one of the hottest topics in generative models. There are three sub-categories including denoising diffusion probabilistic models (DDPMs) \cite{ho2020denoising}, noise-conditioned score networks (NCSNs) \cite{song2019generative}, and stochastic differential equations (SDEs) \cite{song2020score}. Diffusion models operate in two distinct stages: (1) in the forward stage, input data undergoes a gradual transformation through the addition of Gaussian noise over multiple steps, and (2) in the reverse/backward stage, the objective is to reconstruct the original input data from the noise. The noise at each time step is estimated by a neural network, typically based on a U-Net architecture, which is a critical component of the diffusion model's learning process. 

On the basis of the fundamental unconditional diffusion model, many approaches have introduced guidance into the generation process to achieve better results. For example, Guided Diffusion \cite{dhariwal2021diffusion} introduces a classification network into the reverse process and calculates gradients based on the cross-entropy loss to guide the generation sampling at each step. In Semantic Guidance Diffusion \cite{liu2023more}, category guidance is further extended to include guidance based on reference images and text, allowing for more versatile editing effects. Collaborative Diffusion \cite{huang2023collaborative} provides a simple and effective method for the collaboration between different diffusion models, allowing each model to leverage its expertise and achieve high-quality multi-modal control in human face generation and editing. When performing face operations based on diffusion models, it's common to use face representations or identity embeddings as control conditions to guide the generation process.

\subsection{Privacy Theory}
\label{sec:privacy-theory}
\subsubsection{$k$-Anonymity}
$k$-anonymity theory \cite{k-anonymity} contains two-step operations of suppression and generalization of datasets. The theory is based on the assumption of \textit{quasi-identifiers}, that is, the data holder can identify the attributes that may appear in the external data, requiring that each record indistinguishable from $k-1$ other records. \textit{Quasi-identifiers} refer to the attribute set that can be connected with external tables to identify individuals. For example, face data can be associated with identity information through face attributes, and \textit{quasi identifier} can be defined as face attributes and semantic attributes can be defined as facial features or identity features.

$k$-anonymity theory ensures that the attacker cannot determine whether a record is in the data set, nor can it judge whether a specific object has sensitive attributes. If a data set satisfies $k$-anonymity, with only quasi-identifiers of one individual known, the true record can only be chosen with a probability of $1/k$, so that the attacker cannot establish an accurate corresponding relationship with a specific user. This theory provides a certain privacy guarantee but also has some limitations. When background knowledge is available to attackers, homogeneity attack \cite{li2006t} and background knowledge attack \cite{l-diversity} become more effective.


\subsubsection{Differential Privacy}
Differential Privacy originally targeted at differential attacks to protect users' privacy. It aims to provide a way to maximize the accuracy when querying from statistical databases and minimize the possibility of identifying their records \cite{dwork2008differential}.

\textit{Definition.} A randomized method $\mathcal{M}$ meets $(\varepsilon, \delta)$-differential privacy for any pair of neighboring datasets $D$ and $D^{'}$ and for every possible set of outcome $S$, when $\mathcal{M}$ satisfies,
\begin{equation} 
\label{equ:differential-privacy}
\operatorname{Pr}[\mathcal{M}(D) \in S] \leq \exp (\epsilon) \cdot \operatorname{Pr}\left[\mathcal{M}\left(D^{\prime}\right) \in S\right]+\delta, \end{equation}
where $\epsilon$ refers to the privacy budget. A differential privacy mechanism is usually realized by adding random noise. According to different noise distributions, it can be further divided into Laplace mechanism, Exponential mechanism, Gaussian mechanism and so on. 

Differential privacy has widely applied in real-world organizations and acted as a means of legal definitions of privacy such as FERPA \cite{nissim2017bridging} and GDPR \cite{cummings2018role}. Differential privacy has also been applied to machine learning, named differential private machine learning, which aims to train models with formal guarantees implemented by randomizing the training process like adding noise to the gradient. The advanced interaction between differential privacy and machine learning has been discussed in \cite{ji2014differential,abadi2016deep,sarwate2013signal}.

\section{Existing Methods}
\label{sec:methods}

\subsection{Algorithmic Taxonomy}

\begin{figure*}[tb]
    \centering
    \includegraphics[width=0.6\linewidth]{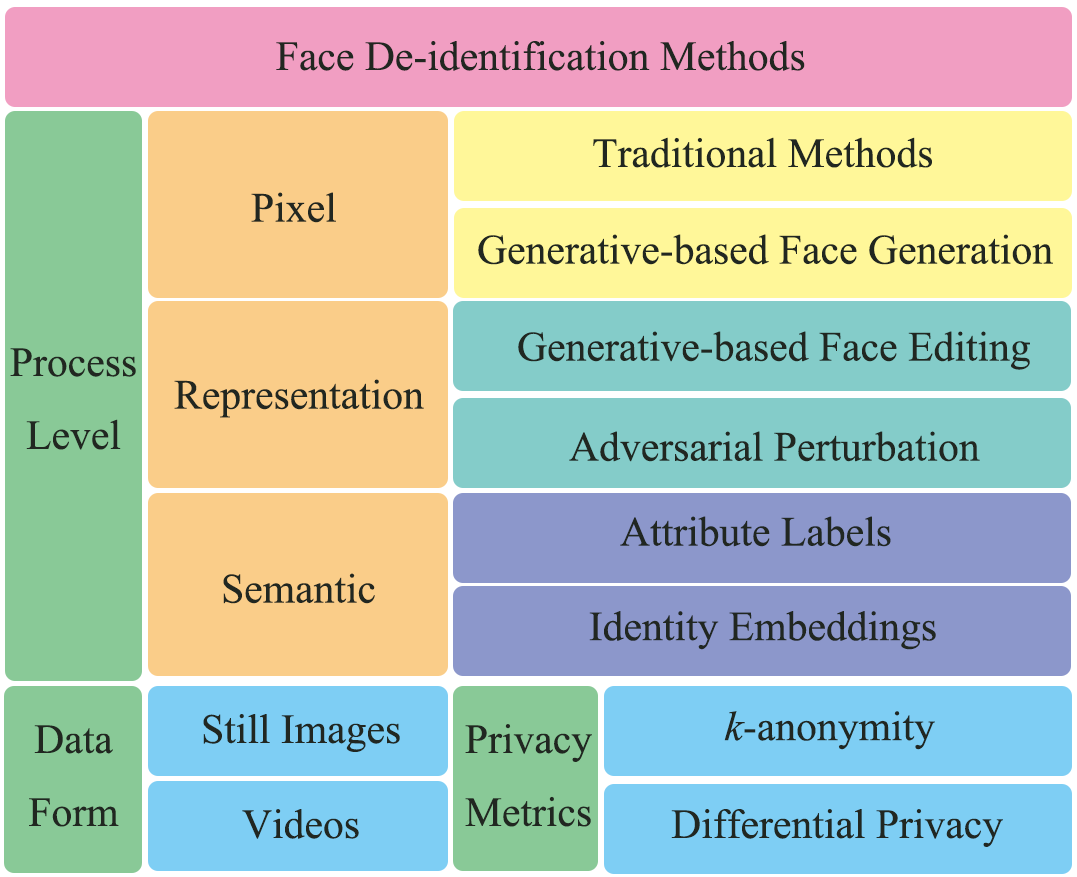}
    \caption{Taxonomy of face de-identification methods. We group existing methods according to process level and then discuss de-identification in videos. In addition, several de-identification methods are combined with privacy theory, like $k$-anonymity and differential privacy, to provide privacy guarantees.
    }
    \label{fig:taxonomy}
\end{figure*}

\begin{figure*}[tb]
    \centering
    \includegraphics[width=0.8\linewidth]{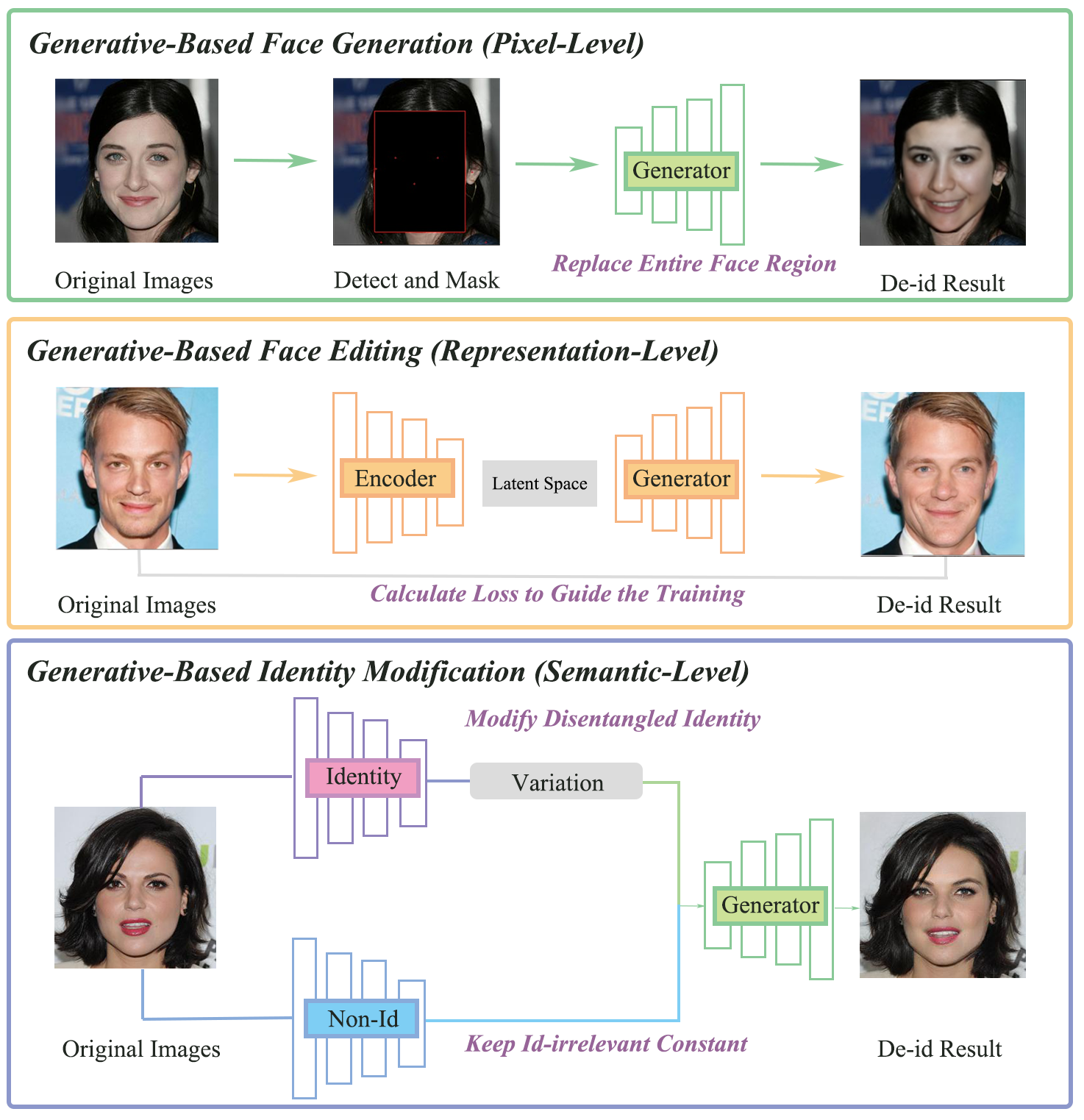}
    \caption{Different applications of Generative Models in de-identification algorithms, where the results are in turn generated by DeepPrivacy \cite{deepprivacy}, SF-GAN \cite{li2021sf} and IdentityDP \cite{wen2022identitydp}.}
    \label{fig:de-id-gan}
\end{figure*}

We propose an algorithmic taxonomy for facial de-identification methods in this survey. They are grouped according to at what level (during an image generation or classification process) and under which data representation form anonymization is addressed. We classify existing methods into three main categories: pixel-level, representation-level and semantic-level. This paper also discussed de-identification in videos and how de-identification algorithms combined with privacy metrics. The full taxonomy is shown in Fig.~\ref{fig:taxonomy}. Especially, generative models, such as Generative Adversarial Networks (GANs) and Diffusion models, are widely used in de-identification methods of deep learning at various levels for better performance, as shown in Fig.~\ref{fig:de-id-gan}.

\textbf{Pixel-level.} Pixel-level methods — as its name suggests — perform facial anonymization at the atomic level, i.e. raw pixels of face images, which could be further divided into three sub-categories. 

First, traditional face de-identification techniques focused on still face images, for instance, black box masking, blurring or pixelation of the facial area after detection and localization.

Second, $k$-Same family introduce $k$-anonymity to facial privacy enhancement. The original face is replaced (hence anonymized) with a synthesized surrogate face, computed through pixel-level averaging. Additional mechanisms \cite{k-same-m,k-furthest,k-same-select,meng2014retaining} to ensure facial attribute similarity or other data and to further enforce identity change are added into developed variants of $k$-Same based methods \cite{k-same}.

Inspired by recent advancements in generative models, especially GANs, the third line of pixel-level de-identification methods revisits the face inpainting or face synthesis task, this time using neural networks. The general idea is to preprocess the input image (usually mask or crop out the identity-revealing facial area), condition the generator on the preprocessed image which generally contains only non-sensitive biometric information (such as background, hairstyles etc.), and leverage a generative network (in most cases conditional GANs) to reconstruct the facial area at pixel-level.

Traditional methods and pixel-level $k$-Same based algorithms are not the focus of this survey. For more detailed information on these methods, readers could refer to other surveys \cite{gross2009face, ribaric2015overview, ribaric2016identification} in the literature.

\textbf{Representation-level.} Representation-level methods perform de-identification directly on or with the help of down-sampled latent representations of a face image in a well-chosen latent space.

Most representation-level strategies rely on generative models, such as Generative Adversarial Networks (GANs), Variational Auto-Encoders (VAEs), and Diffusion models, that are capable of generating synthetic face images with a certain degree of pre-defined characteristics and allow inference over latent features. For Generative models-based techniques, either latent vectors are manipulated directly, 
or indirectly with auxiliary discriminative modules and loss functions that guide the generation process towards de-identification \cite{kuang2021effective, qiu2022novel, li2021sf, wu2019privacy, gu2020password}. 

Another common group of approaches focus on how to fool the face recognition model, which first extracts face embeddings with a well-designed recognition network, and then computes adversarial perturbation to the original face in order to confound face recognition systems. 

We note that the aforementioned $k$-Same based methods could be applied not only on raw image pixels but also combined with deep neural networks for better image quality where high-level image representations in latent spaces are inferred, such as \cite{meden2018k, pan2019k}. 

Alternatively, high-level features of face images could also be modified by adding a noise that follows the differential privacy scheme \cite{liu2021dpimage,chen2021perceptual}. These strategies can help to provide adjustable control on image protection and privacy guarantees.

\textbf{Semantic-level.} Semantic-level methods extract semantic features of a given face image, and achieve privacy enhancement or de-identification by performing obfuscation on certain extracted semantic features. They could be further divided into the following two sub-categories, attribute-level and identity-level.

Attribute-level methods: these methods first leverage some pre-trained networks to obtain a set of semantic attributes (gender, age, glasses, etc.) of a face image. Subsequently, a subset of the extracted features (whose obfuscation leads to the greatest amount of privacy enhancement, or to say the most privacy-revealing subset) is selected according to traditional privacy protection theories and edited towards obfuscation. These methods \cite{anonymousnet,9828699} by design parameterize the de-identification process with respect to various appearance-related characteristics.

Identity-level methods: while attribute-level methods do not explore the relationship between facial attributes and identity, identity-level methods \cite{wen2022identitydp,cao2021personalized,jeong2021ficgan} seek to disentangle the identity embedding from an extracted latent representation and modify or obfuscate only the former. Different from attribute-level methods, identity-level methods first compute one feature map with entangled information on both facial attributes and identities, explicitly disentangle it, and alter the identity embedding but keep the remaining attribute embedding unchanged, which contributes to improving visual similarity and image utility.

Using our level-based classification framework, we now present a comprehensive comparison of the surveyed face de-identification methods in Table~\ref{tab:summary-methods}. In this list, we categorize existing proposals by the following characteristics: 
\begin{itemize}
    \item \textbf{Data Type}: Data type to which the algorithm can be applied, where $I$ stands for still images and $V$ for videos.
    \item \textbf{Controllability}: Whether the degree of image modification, that is, the identity difference between the de-identified result and the original, can be controlled, so as to meet wider application requirements. Among them, although the value of $k$ in $k$-Same family algorithms can be adjusted, it does not correspond to the degree of identity protection directly, so here they are not considered controllable.
    \item \textbf{Diversity}: Whether different results can be generated in the de-identification process.
    \item \textbf{Reversibility}: Whether the image with original identity can be recovered if it has been de-identified.
    \item \textbf{Authenticity}: Whether the identity of the de-identification result corresponds to the real world.
    \item \textbf{Similarity}: Whether to consider preserving the similarity with the original image.
    \item \textbf{Additional Utility}: Whether there are extras designed to preserve specific usability such as attributes immutability, expression consistency, etc.
    \item \textbf{Privacy Guarantees}: Whether it is combined with traditional privacy theory to provide metric guarantees.
\end{itemize}

Especially, due to face de-identification results obtained by \textit{Adversarial Perturbation} keep the same content with original images, where the designed noise cannot be perceived by humans. We use other features to summarize and compare them in Section~\ref{sec:adversarial-pertubation}, so that they aren't listed here for comparison.

\begin{table*}
  \caption{Comparison of the surveyed face de-identification methods. This table summarizes the techniques in terms of process level (Level), data type (DT), controllability (Con.), diversity (Div.), reversibility (Rev.), authenticity (Aut.), similarity (Sim.), additional utility (Add-Utility) and privacy guarantees (Guarantees). Papers in each category are sorted by publication year.}
  \label{tab:summary-methods}
  \footnotesize
  \resizebox{\textwidth}{!}{
  \begin{tabular}{ccccccccccc}
    \toprule
    Level & Methods & Year & DT & Con. & Div. & Rev. & Aut. & Sim. & Add-Utility & Guarantees \\
    \midrule
    \multirow{18}{*}{Pix.} & Blurring & - & I & $\checkmark$ & $\times$ & $\times$ & Real & $\times$ & $\times$ & $\times$ \\
    & Pixelation & - & I & $\checkmark$ & $\times$ & $\times$ & Real & $\times$ & $\times$ & $\times$ \\
    & Masking & - & I & $\times$ & $\times$ & $\times$ & - & $\times$ & $\times$ & $\times$ \\
    & $k$-Same \cite{k-same} & 2005 & I & $\times$ & $\checkmark$ & $\times$ & Fake & $\checkmark$ & $\times$ & $k$-anonymity \\
    & $k$-Same-Select \cite{k-same-select} & 2005 & I & $\times$ & $\checkmark$ & $\times$ & Fake & $\checkmark$ & Expression, Sex & $k$-anonymity \\
    & $k$-Same-M \cite{k-same-m} & 2006 & I & $\times$ & $\checkmark$ & $\times$ & Fake & $\checkmark$ & $\times$ & $k$-anonymity \\
    & $k$-Same-further \cite{k-furthest} & 2014 & I & $\times$ & $\checkmark$ & $\times$ & Fake & $\checkmark$ & $\times$ & $k$-anonymity \\
    & Meng et al. \cite{meng2014retaining} & 2014 & I & $\checkmark$ & $\checkmark$ & $\times$ & Fake & $\checkmark$ & Expression & $k$-anonymity \\
    & Samarzija et al. \cite{samarzija2014approach} & 2014 & I & $\checkmark$ & $\checkmark$ & $\times$ & Fake & $\times$ & Poses & $k$-anonymity \\
    & $k$-Diff-furthest \cite{sun2015distinguishable} & 2015 & I & $\checkmark$ & $\checkmark$ & $\times$ & Fake & $\times$ & $\times$ & $k$-anonymity \\
    & Sun et al. \cite{head-inpainting} & 2018 & I & $\times$ & $\times$ & $\times$ & Fake & $\times$ & $\times$ & $\times$ \\
    & Sun et al. \cite{hybrid} & 2018 & I & $\checkmark$ & $\checkmark$ & $\times$ & Fake & $\times$ & $\times$ & $\times$ \\
    & DP-Pix \cite{fan2018image} & 2018 & I & $\checkmark$ & $\checkmark$ & $\times$ & Real & $\times$ & $\times$ & DP \\     
    & Fan et al. \cite{fan2019practical} & 2019 & I & $\checkmark$ & $\checkmark$ & $\times$ & Real & $\times$ & $\times$ & DP \\
    & DeepPrivacy \cite{deepprivacy} & 2019 & I/V & $\times$ & $\checkmark$ & $\times$ & Fake & $\times$ & $\times$ & $\times$ \\
    & CIAGAN \cite{ciagan} & 2020 & I/V & $\times$ & $\checkmark$ & $\times$ & Real & $\times$ & $\times$ & $\times$ \\
    & JaGAN \cite{balaji2021temporally} & 2021 & V & $\times$ & $\checkmark$ & $\times$ & Fake & $\times$ & $\times$ & $\times$ \\
    & Kim et al. \cite{kim2023method} & 2023 & V & $\checkmark$ & $\checkmark$ & $\times$ & Real & $\times$ & $\times$ & $\times$ \\
    & RID-Twin \cite{mukherjee2024rid} & 2024 & V & $\times$ & $\checkmark$ & $\times$ & Fake & $\times$ & Expression & $\times$ \\
    \hline
    \multirow{18}{*}{Rep.} 
    & Ren et al. \cite{ren2018learning} & 2018 & V & $\times$ & $\times$ & $\times$ & Fake & $\times$ & Action & $\times$ \\
    & $k$-Same-Net \cite{meden2018k} & 2018 & I & $\times$ & $\checkmark$ & $\times$ & Fake & $\checkmark$ & $\times$ & $k$-anonymity \\
    & $k$SS-GAN \cite{pan2019k} & 2019 & I & $\times$ & $\checkmark$ & $\times$ & Fake & $\checkmark$ & $\times$ & $k$-anonymity \\
    & Song et al. \cite{song2019learning} & 2019 & I & $\times$ & $\times$ & $\times$ & Fake & $\checkmark$ & $\times$ & $\times$ \\
    & PP-GAN \cite{wu2019privacy} & 2019 & I & $\times$ & $\times$ & $\times$ & Fake & $\checkmark$ & $\times$ & $\times$ \\
    & PPAPNet \cite{kim2019latent} & 2019 & I & $\times$ & $\times$ & $\times$ & Fake & $\checkmark$ & $\times$ & DP \\
    & Gu et al. \cite{gu2020password} & 2020 & I & $\times$ & $\checkmark$ & $\checkmark$ & Real & $\times$ & $\times$ & $\times$ \\
    & DeIdGAN \cite{kuang2021effective} & 2021 & I & $\times$ & $\checkmark$ & $\times$ & Fake & $\times$ & $\times$ & $k$-anonymity \\
    & DeepBlur \cite{li2021deepblur} & 2021 & I & $\checkmark$ & $\checkmark$ & $\times$ & Fake & $\times$ & $\times$ & $\times$ \\
    & DP-Image \cite{liu2021dpimage} & 2021 & I & $\checkmark$ & $\checkmark$ & $\times$ & Fake & $\checkmark$ & $\times$ & DP \\
    & PI-Net \cite{chen2021perceptual} & 2021 & I & $\checkmark$ & $\checkmark$ & $\times$ & Fake & $\times$ & Attributes & DP \\
    & SF-GAN \cite{li2021sf} & 2021 & I & $\times$ & $\times$ & $\times$ & Fake & $\checkmark$ & Attributes & $\times$ \\
    & QM-VAE \cite{qiu2022novel} & 2022 & I & $\times$ & $\times$ & $\times$ & Fake & $\checkmark$ & Expression & $\times$ \\
    & Yang et al. \cite{yang2022invertible} & 2022 & I & $\times$ & $\checkmark$ & $\checkmark$ & Real & $\times$ & $\times$ & $\times$ \\
    & Zhao et al. \cite{zhao2022private} & 2022 & I & $\checkmark$ & $\checkmark$ & $\times$ & Fake & $\checkmark$ & $\times$ & $\times$ \\
    & Diff-Privacy \cite{he2023diff} & 2023 & I & $\checkmark$ & $\checkmark$ & $\checkmark$ & Fake & $\checkmark$ & $\times$ & $\times$ \\
    & Savic et al. \cite{savic2023identification} & 2023 & V & $\times$ & $\times$ & $\times$ & Fake & $\checkmark$ & $\times$ & $\times$ \\
    & Hanawa et al. \cite{Hanawa2023Face} & 2023 & I & $\times$ & $\checkmark$ & $\times$ & Fake & $\checkmark$ & $\times$ & $\times$ \\
    & Tian et al. \cite{tian2023generative} & 2023 & I & $\times$ & $\times$ & $\times$ & Fake & $\times$ & Race, Age, Gender & $\times$ \\
    \hline
    \multirow{17}{*}{Sem.} 
    & Chi et al. \cite{chi2015face} & 2015 & I & $\times$ & $\checkmark$ & $\times$ & Fake & $\checkmark$ & $\times$ & $k$-anonymity \\
    & Meden et al. \cite{meden2017face} & 2017 & I/V & $\times$ & $\checkmark$ & $\times$ & Fake & $\times$ & Expression & $k$-anonymity \\
    & AnonymousNet \cite{anonymousnet} & 2019 & I & $\times$ & $\checkmark$ & $\times$ & Fake & $\times$ & $\times$ & $t$-closeness \\     
    & CleanIR \cite{cho2020cleanir} & 2020 & I & $\checkmark$ & $\checkmark$ & $\times$ & Fake & $\checkmark$ & $\times$ & $\times$ \\
    & Wen et al. \cite{wen2020hybrid} & 2020 & I & $\checkmark$ & $\checkmark$ & $\times$ & Fake & $\checkmark$ & $\times$ & $\times$ \\
    & CFA-Net \cite{ma2021cfa} & 2021 & I & $\checkmark$ & $\checkmark$ & $\times$ & Fake & $\checkmark$ & $\times$ & $\times$ \\
    & UU-Net \cite{proencca2021uu} & 2021 & V & $\times$ & $\checkmark$ & $\checkmark$ & Fake & $\times$ & $\times$ & $\times$ \\
    & Cao et al. \cite{cao2021personalized} & 2021 & I & $\checkmark$ & $\checkmark$ & $\checkmark$ & Fake & $\checkmark$ & $\times$ & $\times$ \\
    & Chuanlu et al. \cite{chuanlu2021utility} & 2021 & I & $\times$ & $\checkmark$ & $\times$ & Fake & $\times$ & Expression & $k$-anonymity \\
    & Yang et al. \cite{yang2021systematical} & 2021 & I & $\times$ & $\times$ & $\times$ & Fake & $\checkmark$ & $\times$ & $\times$ \\
    & FICGAN \cite{jeong2021ficgan} & 2021 & I & $\times$ & $\checkmark$ & $\times$ & Fake/Real & $\checkmark$ & Attributes & $k$-anonymity \\
    & Croft el al. \cite{croft2021obfuscation} & 2021 & I & $\checkmark$ & $\checkmark$ & $\times$ & Fake & $\checkmark$ & $\times$ & DP \\
    & Cao et al. \cite{9828699} & 2022 & I & $\times$ & $\checkmark$ & $\times$ & Fake & $\checkmark$ & $\times$ & $k$-anonymity, DP \\
    & IdentityDP \cite{wen2022identitydp} & 2022 & I & $\checkmark$ & $\checkmark$ & $\times$ & Fake & $\checkmark$ & $\times$ & DP \\
    & IdentityMask \cite{wen2022identitymask} & 2022 & V & $\checkmark$ & $\checkmark$ & $\checkmark$ & Fake & $\checkmark$ & $\times$ & $\times$ \\
    & Cao et al. \cite{cao2024achieving} & 2023 & I & $\times$ & $\times$ & $\times$ & Fake & $\checkmark$ & $\times$ & $\times$ \\
    & IDeudemon \cite{wen2023divide} & 2023 & I & $\checkmark$ & $\checkmark$ & $\times$ & Fake & $\checkmark$ & $\times$ & $\times$ \\
    & Zhu et al. \cite{Zhu2024IdentityConsistentVD} & 2024 & I/V & $\checkmark$ & $\checkmark$ & $\times$ & Fake & $\checkmark$ & $\times$ & $\times$ \\
  \bottomrule
\end{tabular}}
\end{table*}

\subsection{Pixel-Level}
\subsubsection{Traditional Methods}
\label{sec:traditional-methods}
\textbf{Methods:} In traditional computer vision, privacy protection mainly obfuscates face privacy-sensitive information directly. For example, blurring refers to smoothing the face area and each pixel is replaced by a weighted average of its neighborhood, pixelation reduces the resolution of the face area by dividing the image into a certain range of cells and setting all pixels in each cell as common, masking means to use a black rectangle to occlude the entire face area or key parts, and another way is adding Gaussian or other noise to the pixels. 

On this basis, some methods introduce deep generation models into face blurring. DartBlur \cite{Jiang2023DartBlur} uses the DNN architecture to generate blurred faces, which aims to reduce the artifacts introduced in de-identified results that are harmful to the utility of downstream tasks. The results prove a good balance between accessibility, review convenience and detection of artifact suppression.

\textbf{Discussion:} Though traditional methods have been widely used because of simplicity and ease of operation, it is worth noting that they achieve de-identification by unselectively obfuscating all biometric features of the human face in question, making it hardly reusable in standard computer vision pipelines. In addition, they have limited effectiveness in protecting privacy and are vulnerable to attacks. The study \cite{mcpherson2016defeating} shows that face recognition models based on deep learning technology, especially convolutional neural networks, can still identify people treated with these traditional techniques with high accuracy. What's worse, obfuscation-based protection methods can greatly impair image quality, where visual effects are also greatly reduced since these perturbations can be significantly perceived by humans.

\subsubsection{Generative Model-based Face Generation}
\label{sec:GAN-based-face-generation}
Due to the powerful face generation capability of GAN, it has been introduced to face de-identification. The core idea of achieving privacy protection through face synthesis or inpainting is that if partial or complete face information is used in training the generator, it may bring the privacy leakage risk, so this type proposed to de-identify by reconstructing the face region at pixel-level completely.

\textbf{Methods:} Sun et al. \cite{head-inpainting} proposed an image obfuscation method by head inpainting which can be summarized as the following two stages: (1) if the input image contains the original face area, detect face landmarks with Dlib toolbox while if the original face area is blocked, generate face landmarks with the trained Landmark Generator; (2) generate the head inpainting based on the landmarks and the blackhead or blurhead image with Head Generator. Another identity obfuscation method \cite{hybrid} combined art parametric face synthesis with a deep generative model, where it can manipulate identity explicitly by the parametric part and the generator can help improve the realism of results. It firstly renders the reconstructed face based on 3DMM and replaces the face region. The generator then takes the rendered face and the image of the occluded face area as input, and generates the photo-realism image by adding fine details to match the surrounding image context. 

The most well-known algorithm is DeepPrivacy \cite{deepprivacy}, which proposed a new architecture that can automatically anonymize the face without changing the original posture and image background. 
Based on DeepPrivacy, JaGAN \cite{balaji2021temporally} was designed for face anonymization in videos, masking out faces with black patches and inpainting the missing parts with generated faces by a video generative model enforcing temporal coherence. Compared with the previous work, the innovation of CIAGAN \cite{ciagan} is to control the face anonymity process and guide the synthesis of de-identification results according to the given reference identity. CIAGAN implemented an identity guidance discriminator to drive the generator towards creating images whose representational features are similar to those of a landmark identity, earlier fed into the generator. Thus the anonymization procedure is naturally under full control while diversity in generated identities is simultaneously ensured by carefully selecting landmark identities. CIAGAN took the landmark as input to avoid the leakage of original facial information and the generator conditioned on the face shape can keep the posture information, which also applied the masked background to ensure there are no background changes that may interfere with the detection or tracking. 

To solve the artifacts caused by the difference of face angles, ATFS \cite{Tsai2023ATFSAD} proposes an efficient framework for improved angle transformation and face replacement. In the framework, TransNet and SWGAN are proposed based on generative adversarial networks, where TransNet combines the predicted facial keypoints to locate the target's facial features when reconstructing, and generates multi-angle transformed images to achieve angle transformation, and SWGAN is designed to extract facial features and replace the face region of the input image with the target to realize face de-identification.

\textbf{Discussion:}
De-identification by face inpainting or synthesis can achieve the highest privacy level because they do not require access to the original image and can guarantee the deletion of any privacy-sensitive information. However, they also have the disadvantage that they cannot retain the appearance similarity with the original image. A significant shortcoming is that all information contained in the face region will be removed even if the part is not related to identity. It is hoped to achieve identity protection through minor modification, so that the de-identification results can retain higher utility.

\subsubsection{$k$-Same}
\textbf{Methods:} $k$-Same family algorithms are designed based on $k$-anonymity \cite{k-anonymity} theory. Newton et al. \cite{k-same} proposed to divide the face images in the dataset into $k$ classes according to facial features and calculate the average face in each class to limit the recognition success rate to $1/k$. 
Both $k$-Same-pixels algorithm selecting images for averaging based on Euclidean distance and $k$-Same-eigen algorithm selecting images on Principal Component Analysis coefficient space failed to obtain an average image with good visual effect.

The variants of $k$-Same family aimed at improving the utility and quality of the results. $k$-Same-Select \cite{k-same-select} aimed at preserving facial attributes like gender or expression, which divided the dataset into mutually exclusive subsets according to the data utility function, and apply $k$-Same to different subsets independently. $k$-Same-M \cite{k-same-m} tried to remove unpleasant artifacts because of misalignment by Active Appearance Model (AAM). Another multi-factor framework for face de-identification that unifies linear, bilinear and quadratic models was proposed in \cite{gross2008semi} for better data utility. $k$-Same-further \cite{k-furthest} pointed out that the previous methods target minimizing the identity change through $k$-Same-closest and proposed to select the farthest image cluster to maximize the identity change, no matter how to select $k$ value, it can be perfect privacy protection. Additionally, $k$-Same-further-FET \cite{meng2014retaining} combined $k$-Same-furthest for privacy protection and model-based Facial Expression Transfer for data utility. $k$-Diff-furthest \cite{sun2015distinguishable} adopted the same iterative process of $k$-Same-furthest, where two clusters with size $k$ would be generated in each iteration and their centroids would be swapped. For the problem of poor diversity of the $k$-Same algorithm, $k$-Diff-further could generate a unique de-identified result for each face in a cluster. 

Considering the faces captured in videos have various poses, a de-identification method accommodated for still images with large-angle was proposed in \cite{samarzija2014approach}, which provided the basis for the naturalness of video de-identification. Similar to the ideas of $k$-Same-Select \cite{k-same-select}, face images were grouped in accord with poses, so that the shape could remain in the process of model fitting while the texture would be changed. It also applied $q$-far based on the Euclidean distance to ensure that the appearance distance is far enough.

\textbf{Discussion:} Although the $k$-Same family algorithms are important parts in de-identification which realize the combination of image privacy protection and traditional theories. They also have the following limitations: (1) Because $k$-anonymity theory is vulnerable to background knowledge attack, where the attacker can also reduce the protection effectiveness through the prior knowledge of sensitive information. A more detailed theoretical discussion on the susceptibilities of background knowledge was present in \cite{croft2021obfuscation}. (2) The $k$-Same algorithms assume that the original face image set must be person specific and each object appears only once in dataset, so that the existence of multiple images from the same object or images with similar biometrics will reduce privacy level in a specific practice. In addition, applying the $k$-Same algorithm at pixel level requires images to be strictly cropped out of the face area and aligned in order to reduce artifacts. 

\subsubsection{Differential Privacy}
\textbf{Methods:} 
Fan et al. \cite{fan2018image} first attempted to provide differential privacy guarantees for multimedia data and proposed $m$-neighborhood to protect sensitive information represented by $m$ pixels at most. In addition, a pixelization-based method with grid cells was designed for better privacy-utility balance, and experiments proved that differentially private pixelization could reduce privacy risk significantly. However, applying pixelation to the entire image resulted in high utility loss and the results would be unrecognizable as human faces. For better image quality, another relaxed privacy model in \cite{fan2019practical} developed an efficient mechanism in which Singular Value Decomposition (SVD) was empolyed, and generated visually similar images with the same singular matrices but different singular values. Although image quality could be improved compared with the former, the perceptual information captured by singular value decomposition is limited, and the results generated by perturbing perceptual features of $i$ highest singular values are still far from meeting the requirements of preserving image utility and photo-realism. Additionally, the interactive framework for face obfuscation under the rigorous notion of differential privacy, including DP-Pix \cite{fan2018image} and DP-SVD \cite{fan2019practical} was designed in DP-Shield \cite{saleem2022dp}. 

\textbf{Discussion:} Most methods of applying differential privacy at the pixel-level are to combine it with traditional de-identification algorithms (like pixelation) to provide metric privacy guarantees. Due to the rough processing form, they often failed to generate high-quality results.

\subsection{Representation-Level}
\subsubsection{Generative Models-based Editing}
\label{modifiedGANs}
Another line of face de-identification work seeks to exploit GANs' ability to perform the image-to-image transformation. Seminal advances on GANs have provided an inspiring framework, and in order to adapt it to face de-identification scenarios, some methods attempt to modify the initial architecture of deep generative networks, in particular by attaching auxiliary modules and additional optimizing loss functions. 

\textbf{Method:}
SF-GAN \cite{li2021sf} adds a unique unit to ensure that the data distribution of the synthesized face is different from anyone in the datasets and balances the trade-off between face attribute preservation and privacy protection.
PP-GAN proposed by Wu et al. \cite{wu2019privacy} combines the two principal objectives in face de-identification — de-identify and keep structural similarity — by adding to a basic conditional GAN backbone two additional modules, (1) a verificator that enforces the sampling range away from given samples in the identity embedding space, or to de-identify, and (2) a regulator which constrains the variation of image structure, or to retain structural similarity. Both are explicitly expressed as external losses to compensate for their contradiction in traditional GAN loss. Zhao et al. \cite{zhao2022private} designed an adjustable privacy-related loss in the training process, which indicates the identity distance between results and the origin. A pre-trained styleGAN2 was used as the generator, so that high-quality and high-resolution de-identification results can be obtained. Although different parameters can be set to generate de-identified images with different privacy levels, the parameters lie in the loss function, so the encoder needs to be retrained, where the flexibility of the model would be limited. Hanawa et al. \cite{Hanawa2023Face} introduce a de-identification loss function and embed facial features of other persons into the face images.

Several methods transformed face images into latent space, and generated de-identified results with regard to the modified representations. DeepBlur \cite{li2021deepblur} searched the latent vectors with a feature extractor, then applied a Gaussian filter on them, and finally reconstructed with the blurred representations. Various kernel sizes of Gaussian filters can be selected to obtain different results with desired privacy levels.

Noting the irreversibility and hence limited data accessibility in previous methods, Gu et al. \cite{gu2020password} further proposed a password-based face identity transformer enabling anonymization and de-anonymization. With carefully designed optimizing loss functions, this model could (1) remove face identity information after anonymization, (2) recover the original face when given the correct password, and (3) return a wrong but photo-realistic face with a wrong password. An invertible mask for face privacy was proposed in \cite{yang2022invertible}. The swapped image was generated according to the protected face and replaced face and a well-designed ``mask" would be put on to obtain the masked face and calculate lost information, where the ``mask" is visually indistinguishable. For authorized users, the original image could be recovered by operating a discrete wavelet transform on the masked face and auxiliary information. FaceRSA \cite{Zhang2024FaceRSA} is the first fully facial identity cryptography framework similar to RSA based on StyleGAN, which can control the face identity by user-given password for anonymization and deanonymization. The unrelated attributes can be kept by adding modifications only to the identity-related layers of embeddings in StyleGAN latent space. \textit{Single loss} is used for a single pair of encryption and decryption processes, \textit{Sequential loss} is designed for more complex requirements with multiple pairs of passwords, and \textit{Associated loss} is applied to ensure that equivalent passwords exist during both encryption and decryption.

Based on evolutionary GANs \cite{wang2019evolutionary}, Song et al. \cite{song2019learning} opted for another approach, which dynamically selects an optimal generator that best caters for the de-identification need. Their method differs from the previous ones in that the distance and structural similarity between the original and de-identified faces are introduced during the evaluation and selection process of generators, and thus no supplementary networks are needed.

QM-VAE (Quality Maintenance-Variational AutoEncoder) framework \cite{qiu2022novel} was a multi-input generative model guided by service quality evaluation. In the first step, four sets of de-identified images without private information were generated by four typical methods (blindfold, cartoon, Laplace, and mosaic). Next, these datasets were fed into QM-VAE to calculate whether the utility was consistent with the original data. At the same time, the calculation results were added as a part of the loss function for model training. The service quality was quantified as the difference of attributes in service scenarios, and it contributes to generating reliable faces.

Diff-Privacy \cite{he2023diff} is the first work to apply the diffusion model to privacy protection, addressing two crucial tasks in facial privacy: anonymization and visual identity information hiding. The former aims to prevent machine recognition, while the latter requires ensuring the accuracy of machine recognition. By training the proposed Multi-Scale Image Inversion (MSI) module, a set of conditional embeddings in the SDM format is obtained from the original images. Based on these conditional embeddings, corresponding embedding scheduling strategies are designed to construct different energy functions during the denoising process.

\textbf{Discussion:}
The core principle of Generative Models-based methods is to impose certain constraints by adding external guidance losses (guidance towards target identity or away from source identity, guidance to maintain certain attributes, etc.), together with their accompanying networks (namely verificators, regulators, modifiers, etc.).
Generative Models-based methods generally achieve promising de-identification rates, but they sometimes struggle in applying to faces in the wild with non-frontal head poses or occlusions, and are not evidently applicable to videos \cite{wu2019privacy}. Meanwhile, although they generate realistic de-identified faces, their output could still drastically alter the original faces in terms of visual perception and would, therefore, fail to live up to the possible expectation of retaining the original face appearance. Another shortcoming of Generative Models-based methods is that the quality of generated faces is highly dependent on the implied structure itself, thus general artifacts could be noticed in their outputs \cite{gu2020password}. Taking the basic generative models as a baseline and adding appropriate constraints to improve the network, the purpose of de-identification can be effectively realized. Unfortunately, the degree of privacy protection cannot be adjusted if the model has been trained. Models with different parameters are required to meet different applications, so the flexibility may be poor.

\subsubsection{Adversarial Perturbation}  
\label{sec:adversarial-pertubation}
The existing face de-identification methods based on adversarial examples are mainly aimed at preventing the faces from being recognized by the deep neural network, so as to prevent private them from being used to make false images or videos.

\textbf{Methods:} The latest research on adversarial image perturbations \cite{aip} has shown that it can effectively confuse the face recognition system without introducing significant artifacts, and proposed to use of the game theory framework to achieve the opposite goals for users and recognizers to obtain the privacy guarantee of the user without the countermeasures independent of the recognizers.

A series of methods could generate strong perturbations over the entire facial images. A$^3$GN (Attentional Adversarial Attack Generative Network) \cite{yang2021attacks} introduced face recognition networks as the third player for target-attack, where a conditional variational autoencoder was applied to get latent code and attentional modules were provided to capture feature representation and dependencies. Dong et al. \cite{dong2019efficient} proposed an evolutionary attack method for query-efficient adversarial attacks for the black-box scenario. The target is to model the local geometries of search directions and reduce the dimension. DFANet (Dropout Face Attacking Network) \cite{zhong2020towards} applied dropout layers to the surrogate model in each iteration. Inspired by the I-FGVM \cite{i-fgsm} algorithm, P-FGVM \cite{p-fgvm} has been proposed to generate adversarial de-identified facial images that resemble the original ones in the image spatial domain, which can both protect the privacy and preserve visual facial image quality. AdvFaces \cite{deb2020advfaces} could focus on the set of pixel locations required by face recognition model and only add perturbation to those salient regions like eyebrows and nose. The most advanced method Fawkes \cite{shan2020fawkes} helps users add an invisible ``cloak" to the shared image, which can successfully confuse the face recognition model. CSP${^3}$Adv \cite{pan2023collaborative} proposed a collaborative privacy protection method based on adversarial examples for social networks, which avoids direct image acquisition by the server and the protected images are robust to JPEG compression. Users can construct the gradient perturbation on the lightweight face recognition model, and the perturbation transfer module is trained to learn the corresponding perturbation under the complex face recognition model on the server side. Additionally, the high-frequency noises will be filtered to improve the privacy protection effect after social network compression. Ghafourian et al. \cite{Ghafourian2023Toward} compare the effectiveness of two popular adversarial methods, Basic Iterative Method (BIM) and Iterative Least Likely Class (ILLC) to de-identify. They reach the preliminary conclusion that BIM works better through experiments and transferability between multiple face recognition models.

Patch-based adversarial attacks mean just adding adversarial noises to certain areas. For example, the person can be allowed to evade being recognized when wearing the generated eyeglass in \cite{Accessorize}. AdvHat \cite{komkov2021advhat}  was designed to create a rectangular image stuck on the hat and decrease similarity to ground-truth identity below threshold. Although these algorithms will not affect the naturalness of the generated results, these adversarial patches are perceptible to humans and mostly applied in the white-box setting.

For increased robustness and image quality, Yang et al. \cite{yang2022generating} introduced an attention module to different target classification networks to extract chief features and processed them by feature fusion, which helps to improve the transferability. Almost invisible adversarial examples were generated by perturbing the features of faces calculated by feature fusion matrix.

Some studies combined de-identification with other face editing tasks, where the adversarial noise can be hidden within the editing area to obtain better visual effects. A common example is makeup transformation. Zhu et al. \cite{zhu2019generating} made the first attempt by applying two sub-networks, the makeup transfer sub-network and the adversarial attack sub-network for the white-box face recognition attack. Adv-Makeup \cite{yin2021adv} proposed to add makeup to eye regions to fool face recognition models. They applied a blending method to improve the naturalness of generated results and a fine-grained meta-learning strategy to improve the transferability of the black-box attack. AMT-GAN (Adversarial Makeup Transfer GAN) \cite{hu2022protecting} proposed to generate an antagonistic face image with makeup transferred from the reference image. A new regularization module and a joint training strategy were introduced to coordinate the conflict between adversarial noise and cycle consistency loss in makeup transfer, so as to achieve an ideal balance between image quality and attack success.

\begin{table*}
\centering
  \caption{Comparison of de-identification based on Adversarial Perturbation. Papers are sorted by publication year.}
  \label{tab:adv-methods}
  \begin{small}
  \resizebox{0.8\textwidth}{!}{
  \begin{tabular}{ccccc}
    \toprule
    Methods & Year & Model Knowledge & Attacks Type & Perturbation Region\\
    \midrule
    Sharif et al. \cite{Accessorize} & 2016 & White / Black-Box & Target / Non-target & Eyeglasses \\
    Dong et al. \cite{dong2019efficient} & 2019 & Black-Box & Target & Entire Face\\
    Zhu et al. \cite{zhu2019generating} & 2019 & White / Black-Box & Target / Non-target & Eye (Makeup)\\
    P-FGSM \cite{p-fgvm} & 2019 & White-Box & Target & Entire Face\\
    AdvFaces \cite{deb2020advfaces} & 2020 & Black-Box & Target / Non-target & Eyebrows, Nose\\
    Fawkes \cite{shan2020fawkes} & 2020 & Black-Box & Non-target & Entire Face\\
    DFANet \cite{zhong2020towards} & 2020 & Black-Box & Target & Entire Face\\
    AdvHat \cite{komkov2021advhat} & 2021 & White-Box & Non-target &Hat\\
    A$^3$GN \cite{yang2021attacks} & 2021 & White / Black-Box & Target & Entire Face\\
    Adv-Makeup \cite{yin2021adv} & 2021 & Black-Box & Target & Eye Shadow \\
    AMT-GAN \cite{hu2022protecting} & 2022 & Black-Box & Target & Entire Face (Makeup)\\
    Yang et al. \cite{yang2022generating} & 2022 & White / Black-Box & Non-target & Entire Face\\
    CSP${^3}$Adv \cite{pan2023collaborative} & 2023 & White / Black-box & Target & Entire Face \\
  \bottomrule
\end{tabular}}
\end{small}
\end{table*}

\textbf{Discussion:} Unlike other face de-identification algorithms, the adversarial perturbation techniques aim at disturbing the face recognition model without any modification to the image content. They cannot affect visual observation and identity recognition of humans as they retain the appearance exactly the same as the original image. Therefore, the protection performance and application scenarios of this type of face de-identification method are also different. In addition, it is a practical research that combines adversarial examples with other face modification tasks to make the adversarial noises for de-identification inconspicuous and unnoticeable. 

There are still various issues with existing methods. Compared with white-box attacks, black-box attacks are more useful in real-world scenarios. Like other adversarial attack research, a major challenge is how to improve the generalization and robustness of adversarial examples. On the other hand, there exist obvious artifacts in the de-identification results and the clarity can still be improved.

\subsubsection{$k$-Same}
\textbf{Methods:} 
Chang et al. \cite{Chang2023Machine} focused on the requirement of machine learning data utility of face de-identification and proposed to adjust the data class of $k$-anonymity by feature important (FI) and margin preservation (MP), where margin preservation can not only anonymity the data but also can retain the original boundaries.

With the development of deep learning, some algorithms combine $k$-same theory with the deep generation network in order to solve the problem of poor visual perceptions. $k$-Same-Net \cite{meden2018k} generated anonymity face by generative networks instead of image or model-parameter averaging. However, $k$-Same-Net can only generate synthetic images with trained identities, which need to be retrained for each new image. $k$-Same-Siamese-GAN \cite{pan2019k} proposed to take advantage of the mixed precision training (MPT) technique for better privacy protection and image quality. In addition, $k$-anonymity theory has been widely applied in many face de-identification algorithms. Anonfaces \cite{le2020anonfaces} extracted low-dimensional features of face images and performed fixed-size clustering analysis while minimizing information loss. The trade-off between privacy and utility can be adjusted by fine-tuning the weight in identity fusion or selecting an appropriate $k$ value. Compared with $k$-Same-Net, Anonfaces could generate any de-identification result with an unknown identity. 

An identity-adversarial discriminator is leveraged to guide the generator away from synthesizing the original identity in DeIdGAN \cite{kuang2021effective} and pre-defined sensitive identities. DeIdGAN remove the original identity information by anonymizing both shape and style, where the de-identified results are generated with $k$-anonymized ASM (Anonymous Semantic Mask) and ASI (Anonymous Style Image).

\textbf{Discussion:} The de-identification results lack naturalness, diversity and realism, even though there is a trend of combining with deep learning to improve image quality. Each method based on $k$-Same has a common disadvantage of failing to provide unique de-identified outputs for various images \cite{qiu2022novel}.

\subsubsection{Differential Privacy}
\textbf{Methods:} It is a common idea to expand the traditional differential privacy protection theory when applying it to high-level features. DP-Image \cite{liu2021dpimage} firstly introduced a novel notion of image-aware differential privacy, which redefined traditional DP and local differential privacy (LDP) \cite{duchi2013local}. DP-Image provided provable and adjustable privacy levels to achieve image protection through adding noise to image features combined with advanced GAN techniques. The pSp framework \cite{richardson2021encoding} performed as feature extraction to map input image into latent space, the noise generator used to inject noise onto the feature, and StyleGAN2 \cite{karras2020analyzing} acted as the image reconstruction module to generate protected results with perturbed features.

Focusing on that applying DP with Laplace noise may lead to unacceptable artifacts, another method named PI-Net (Perceptual Indistinguishability-Net) \cite{chen2021perceptual} was proposed to study how differential privacy should be applied for face image obfuscation. Motivated by metric privacy \cite{chatzikokolakis2013broadening}, image adjacency was defined as the difference between high-level features. PI-Net also provided a clear semantic definition of latent space which can be captured for privacy analysis. Based on the definition of perceptual distance and adjacent images, $\epsilon$-PI was proposed to provide a privacy guarantee when injecting noises to obfuscate images.

The PPAPNet \cite{kim2019latent} (Privacy-Preserving Adversarial Protector Network) proposed to add optimized noise at the latent space level, consisting of protector, critic and attacker. Protector was designed as an encoder-decoder network for obtaining latent codes, and another encoder-decoder network, attacker, tried to prevent the anonymized results from model inversion attacks \cite{fredrikson2015model}. Especially, a noise amplifier inside the protector was trained to generate optimal noise for latent representations of each dimension, which contributed to both image realism and privacy attack immunity.

\textbf{Discussion:} Lack of formal privacy guarantee inspired many scholars to introduce differential privacy, a provable guarantee of privacy, in face de-identification process. Due to the difference in the data form of face images, how to combine deep generative network and differential privacy theory and how to improve traditional DP to make it more suitable for face de-identification tasks are major problems worthy of attention in the future.

\subsection{Semantic-Level}
\subsubsection{Facial Attribute Labels}
\label{sec:attributes}
Some researchers regard facial attribute labels as privacy-sensitive information and realize image identity protection based on attributes indistinguishability. This type of algorithm is usually combined with the face attribute editing model in the implementation process. On the contrary, some algorithms aim at de-identification with attribute preservation to ensure the utility in attribute classification tasks, which will be discussed in Section~\ref{sec:advanced-utility}.

\textbf{Method:} 
AnonymousNet \cite{anonymousnet} proposed a de-identification framework based on attribute editing and designed privacy-preserving attribute selection (PPAS) algorithm based on traditional privacy protection theory to make the attribute distribution of the protected face close to that in real life, which is the first to prove that face privacy is measurable and provide privacy guarantees. AnonymousNet consists of four steps: facial feature extraction, semantic-based attribute obfuscation, de-identified face generation based on StarGAN \cite{choi2018stargan}, and adversarial perturbation. Further considering minor changes to protected images while providing metric privacy, a differential privacy $k$-anonymity algorithm \cite{9828699} was proposed to calculate the obfuscated attributes as the de-identified target attributes. 
In UU-Net \cite{proencca2021uu}, four attributes \textit{ID, gender, ethnicity, hairstyle} labels were taken into consideration, where \textit{ID} is always selected to change for de-identification. UU-Net can have full control over the properties of anonymized faces in poor-resolution video streams.


\textbf{Discussion:}
Since facial attribute tags can be represented by binary discreteness, most of these algorithms can be combined with traditional privacy protection theories like $k$-anonymity \cite{k-anonymity}, thereby enhancing interpretability. However, most algorithms do not further consider the relationship between face attributes and identity. Most methods focus on the attributes which have a great impact on the appearance rather than the attributes with higher privacy level. For example, we think external features like hair color have weak privacy to be protected but have a great impact on appearance. 
Identity preserving face privacy protection is a related topic, which tries to anonymize facial attributes but preserve the recognition utility. The soft biometric privacy was discussed in \cite{othman2014privacy,mirjalili2017soft} and some researches targeted at multi-attribute privacy, such as PrivacyNet \cite{mirjalili2020privacynet} and SensitiveNet \cite{vera2021sensitivenets}. Li et al. \cite{li2021identity} proposed a face anonymization method that is composed of an identity-aware region discovery module and an identity-aware face confusion module. Different from removing identity information, these methods could obfuscate visual appearance while retaining identity discriminability. There are also some studies that use attribute inference as the goal of privacy. For example, InfoScrub \cite{wang2021infoscrub} employed a bi-directional discriminator for attribute inversion in the first step, and performed attribute obfuscation by maximizing entropy on inferences over targeted privacy attributes.
They are also interesting research fields, and the impact of identity and attribute labels on face privacy is also worth studying.

\subsubsection{Identity Embeddings}
\label{sec:identity-representation}
Disentanglement is based on the assumption that each component of the representation learned by neural networks is independent of each other. It aims at transforming the entangled data into another latent space that can be separated. How to disentangle identity and attributes has also received extensive attention in face swapping and face de-identification tasks.
In this section, we reviewed face swapping based on disentanglement, Generative Models-based identity modification for de-identification, and combining $k$-anonymity or differential privacy to identity encryption. 

Face swapping refers to transferring the identity features from the source image to the target while keeping other attributes, such as posture, expression, and lighting conditions invariant.
An important difference is that the de-identified results may not correspond to a real identity but the swapped results should keep consistent with the source in recognition.
In order to avoid unnecessary face modification, some Generative Models-Based algorithms have been proposed to only modify the disentangled identity representation. Since the identity can be represented as a vector, it is also conducive to the further integration of privacy protection theory.

The basic framework can be summarized as an encoder-decoder structure, where the encoder can be divided into identity encoder and attribute encoder \cite{nitzan2020face,li2022learning, li2019faceshifter, wen2020hybrid,cho2020cleanir}, and the decoder acts as a generator. Due to the development of face recognition technology \cite{arcface,cosface,facenet}, identity features can be represented as a vector. Relatively, attribute features should include all other identity-irrelevant information, so that the network structure and the representation will be more complex. Most methods applied a pre-trained face recognition model as the identity encoder to obtain identity embedding to guide the training of the attribute encoder and generator.

\textbf{Face Swapping.}
Better design of network structure or auxiliary module is a common improvement idea. FaceShifter \cite{li2019faceshifter} designed the adaptive embedding integration network (AEI-Net), where an adaptive attentional denormalization generator can achieve adaptive integration of identity and attributes. Additionally, a novel heuristic error acknowledging refinement network (HEAR-Net) is applied in the second stage to address face occlusions. ZFSNet (Zero-shot Face Swapping Network) \cite{li2022zero} proposed an additional de-identification module and an attention spatial style modulation to blend features. A classifier designed on a multi-layer perception network is employed in the de-identification module to obtain identity-related features. FaceSwapper \cite{li2022learning} is another end-to-end framework proposed for one-shot progressive face swapping. There are two major parts, the disentangled representation module and the semantic-guided fusion module. 
Some studies have improved the connotation of identity. For example, HifiFace \cite{wang2021hififace} proposed 3D shape-aware identity based on the idea that face shape is the feature corresponding to identity. HifiFace introduced a pre-trained 3D face reconstruction model \cite{deng2019accurate} as a shape feature encoder combined with a state-of-the-art face recognition model to get the identity information including geometric structure.

Some methods directly use the pre-trained StyleGAN \cite{karras2019style} as the generator for high-quality image generation. Nitzan et al. \cite{nitzan2020face} proposed to learn a mapping network to map disentangled face representations $z$ into $W$ space. Xu et al. \cite{xu2022high} disentangled the latent semantics of StyleGAN by utilizing the progressive nature of the generator and deriving structure and appearance from different layers.

There are also some approaches that take advantage of the diffusion model for the face-swapping task. DiffFace \cite{kim2022diffface} proposed an ID Conditional DDPM, which can generate face images with the desired identity. Some facial expert models are introduced in the sampling process to preserve target attributes. DiffSwap \cite{zhao2023diffswap} introduces a high-fidelity controllable face-swapping approach, treating it as a conditional restoration task, executed by a powerful diffusion model guided by facial identity and landmarks. Additionally, a midpoint estimation method is proposed to efficiently recover reasonable diffusion results for the swapped faces in just two steps, enabling the introduction of identity constraints to enhance the quality of face swapping.

\textbf{Generative Models-Based Identity Modification.} Many de-identification algorithms of this type include two steps, disentanglement and identity conversion. CleanIR \cite{cho2020cleanir} adopted the VAE architecture with skip connections to split encoded features into identity-related and attributes-related parts, and the Identity Changer module was designed to calculate a new identity for de-identification. Wen et al. \cite{wen2020hybrid} proposed to add adjustable privacy to the individual identity embedding, which achieves the target of generating natural de-identification results and controlling the degree of privacy protection according to application scenarios. ITNet \cite{Wang2023Identifiable} randomly assigns a unique virtual identity mask added to the disentangled identity. In addition, the identity mask is optimized in de-identification using the identities extracted from the user's $n$ images. CFA-Net (Controllable Face Anonymization Network) \cite{ma2021cfa} disentangled face identity from other contents, and projected identity onto spherical space for manipulation. The anonymized results changed in two dimensions, anonymous extent and anonymous identity variety. Due to the identity distance can be expressed as cosine similarity between features calculated by face recognition model, when the angle of identity representations in spherical space is larger than a pre-set threshold, the identity distance will exceed the threshold of face recognition model, achieving the target of de-identification. A personalized and invertible face de-identification method was proposed in \cite{cao2021personalized}, where the encryption process in hyperspherical space is independent of a deep generative network. Various de-identification results can be generated with the user-specific password, the degree of identity variation can be controlled by privacy level parameters and the original identity can be restored with the correct password. Similarly, MRDD-FID \cite{Xiao2024ManipulableRA} proposes a manipulable, reversible, and diverse de-identification framework that enables individuals to only modify identity-related representations. Random passwords are used for identity variation to ensure the randomness of the identity modification. MRDD-FID applied StyleGAN2 as a high-quality face generator and injected identity features by the weight modulation/demodulation layer. The degree and direction of de-identification can be precisely controlled according to the intensity and style specified by the user, without affecting the image quality.

Another systematical face de-identification method \cite{yang2021systematical} proposed to de-id for humans by face swapping and de-id for face recognition model by adversarial perturbations. Similarly, a pre-trained face recognition model is used as the identity encoder and another Inception-V3 network acts as the expression and pose encoder, where the latent vector is obtained by combining the two parts. Additionally, the adversarial vector mapping network was designed to calculate the adversarial latent vector by propagating the gradient back instead of adding adversarial noise to the original images. Finally, the de-identified results could be generated with the adversarial latent vector by StyleGAN.

Some recent methods introduce more advanced 3D face generation models into face de-identification. IDeudemon \cite{wen2023divide} employs a ``divide and conquer" strategy, gradually preserving identity and utility while maintaining good interpretability. In Step I, the 3D reconstruction ID codes calculated by the NeRF model are blurred to protect identity. In the second step, a face restoration GAN with adjusted losses has been introduced for visual similarity assistance. Noticed that 2D generation models are difficult to produce identity-consistent results for multiple views, Cao et al. \cite{cao2024achieving} focus on identity disentanglement in the latest 3D-aware face generation models and propose a comprehensive framework that can be applied to single-view, multi-view and a group images of the same identity.

\textbf{$k$-Same.} Meden et al. \cite{meden2017face} exploited GNNs to generate surrogate faces, which first extract feature vector by a pre-trained face network and select $k$ most similar identities in the gallery for de-identified results generation. Chi et al. \cite{chi2015face} proposed to generate ``average face" by suppressing facial identity-preserving (FIP) features, where FIP features can reduce intra-identity variances while maintaining inter-identity distinctions. At the same time, FIP features are invariant to pose and illumination, and the de-identification results are generated by replacing that of the original with the averaged FIP features of $k$ individuals. Based on the hypothesis that face appearance contains identity factor and utility factor (expression, sex, etc.), Chuanlu et al. \cite{chuanlu2021utility} proposed to separate the multi-dimensional representation with regard to identity and expression with Information Gain (IG) feature selection method. In the de-identification process, $k$-anonymity procedure is just applied in the identity subspace while keeping the utility subspace consistent.

FICGAN \cite{jeong2021ficgan} was developed as an autoencoder-based conditional generative model, where the ID and non-ID latent codes disentanglement and layer-wise generator structure ensure the controllability between de-identification and attribute preservation. Two types of contrastive verification loss \cite{kang2020contragan, zhang2019consistency} were introduced for a better identity. Non-ID latent code is inserted into the generator for structural attributes and ID latent code is processed by adaptive instance normalization (AdaIN) \cite{huang2017arbitrary}. FICGAN provided various privacy protection modes including manifold $k$-same algorithm \cite{k-anonymity}, controlling degree, attribute control, example-based control and face swapping. 

\textbf{Differential Privacy.} Differential Privacy has also been introduced into face de-identification methods at the semantic level, especially identity representations. Croft et al. \cite{croft2021obfuscation} proposed the first framework to apply differential privacy to facial identity obfuscation via generative machine learning models. The network took one or more class vectors as input and each identity vector was considered as a specific class to be obfuscated. In order to solve the problem that unknown identities cannot be generated, it is proposed to approximate an unknown identity as a weighted sum of the known identities and formulate the approximation as an optimization problem. IdentityDP \cite{wen2022identitydp} was the first to introduce the rigorously formulated DP theory into face anonymization and applied differential privacy mechanisms to deep neural networks for adjustable privacy control. The framework includes facial representations disentanglement, $\varepsilon$-IdentityDP perturbation and image reconstruction, where various de-identification results can be generated according to different privacy budgets. When adding differential privacy perturbation to the identity representation, those attributes remain unchanged for sharing more similarities with the original.

\textbf{Discussion:} At present, it is considered that adding appropriate disturbance to the disentangled identity embedding is an effective method to preserve the visual similarity as possible, which can obtain visually similar face photos while hiding identity information. How to design the model to learn a more accurate identity representation and how to disentangle the identity and attributes more effectively are the focus of further research.



\subsection{De-identification in Videos}
\label{sec:deid-videos}
Compared to image processing, the processing of video is more complex, including the consistency of adjacent frames, the temporal consistency of identity, the naturalness of background in generated images and so on.

\textbf{Method:}
Enforcing spatio-temporal constraints is important for extending image-based methods to video face de-identification. For CIAGAN \cite{ciagan}, the same framework could be used for both images and videos because its landmark-based representation of the input face ensures pose preservation and rudimentary temporal consistency. In a similar attempt and inspired by Vid2Vid GAN \cite{wang2018video}, Balaji et al. \cite{balaji2021temporally} utilized an initially image-based GAN to anonymize video frames, and coupled it with several mechanisms to guarantee temporal coherence whilst video processing. Notably, a conditional video discriminator, a pre-processing phase, and a ``burn-in" stage, respectively, enforce consistency between consecutive frames, within each sequence and across sequences. FIVA \cite{rosberg2023fiva} is able to maintain the same face anonymization consistently over frames with a suggested identity-tracking and guarantees a strong difference from the original face. 

Gafni et al. \cite{gafni2019live} proposed a novel feed-forward encoder-decoder network to de-identify live videos. The model holds a host of novelties. First, the representation from an existing face recognition network is concatenated to enrich the latent space, thereby augmenting the auto-encoder to be able to feed-forwardly treat new identities unseen during training. Second, a new attractor-repeller perceptual loss is introduced to maintain source expression, pose and lighting conditions, while surrogating the target identity. This is achieved by employing a perceptual loss between the source and generated output on several low-to-medium abstraction layers, while distancing the high abstraction layer perceptual loss between the target and generated output. 

Savic et al. \cite{savic2023identification} designed a semi-adversarial framework which processes an input video by adding unobtrusive perturbations that remove biometric privacy while preserving the rPPG signal and visual appearance. Unlike the previous works that deal with pronounced facial spatial features, they aim at preserving the underlying physiological signals that depend on subtle spatial-temporal features. This means that the applied perturbations need to be temporally consistent and not detrimental to the subtle visual cues that compose the rPPG signal. Therefore, they take interframe dependencies into account when introducing an rPPG preserving constraint, and choose a 3D-CNN as the autoencoder for spatial-temporal modeling. The features are only spatially upsampled and downsampled to not degrade the delicate temporal information.

Some other researchers tackle video anonymization by swapping faces. Among face-swapping models, the most famous one is 
faceswap/deepfake project \cite{deepfake}. It first trains two auto-encoders with a common encoder but different decoders, one on the target face dataset and the other on the source face dataset. Then face swap is achieved by exchanging the two decoders. Zhu et al. \cite{zhu2020deepfakes}, showcased the utility of this method in the video de-identification task and proved that it is key point invariant.

Afterwards, Mukherjee et al. \cite{mukherjee2024rid} proposed RID-Twin, which stands for Re-enacting Inpainting based De-identified-Twin. Specifically, they decouple human identity from motion and re-enact a de-identified twin actor, thereby transferring motion to a de-identified image. They evaluated their pipeline regarding the de-identification level, identity consistency, and expression preservation.

Furthermore, Park et al. \cite{park2024verifiable} introduced a framework that combines face verification-enabled deidentification techniques with face-swapping methods, tailored for video surveillance environments. This method maintains high face recognition performance (98.37\%) across various facial recognition models while achieving effective de-identification.

Most methods in the literature require facial landmarks to perform video de-identification. However, Proença \cite{proencca2021uu} proposed UU-Net, a landmarks-free and reversible face de-identification method for video surveillance data. UU-Net is composed of two sequential modules, one that receives and de-identifies the raw stream, and the other that reconstructs, disclosing the original identities. Joint optimization of these two entities intrinsically implies shared knowledge on encoded features, assuring proper reconstruction, hence reversibility. Also, he keep full control over the facial attributes that are preserved/changed between the raw and de-identified streams. This is achieved by inferring a pre-trained multi-label CNN classifier to estimate agreeing/disagreeing labels (ID, gender, etc.) between images, whose pairwise responses are later used to constrain the properties of the de-identified elements, as well as temporal consistency in the same sequence.

Zhu et al. \cite{Zhu2024IdentityConsistentVD} introduced an innovative framework for video face de-identification that leverages diffusion autoencoders to disentangle identity and motion features from videos. Unlike existing GAN-based methods, this model offers a robust approach that excels in both reconstruction and de-identification tasks due to its foundation in diffusion models.

While previous methods emphasize designing novel networks to process face videos frame by frame, Wen et al. \cite{wen2022identitymask} was the first to introduce motion flow into video de-identification tasks to avoid per-frame evaluation. Their method first infers a motion flow generator from \cite{siarohin2019first} to compute the relative dense motion flow between pairs of adjacent original frames and runs the de-identification only on the first frame. The de-identified video will be obtained based on the anonymous first frame via the relatively dense motion flow.

Meanwhile, because videos produced through live streaming can cause problems such as infringement of others’ portrait rights and exposure of trademarks, Kim et al. \cite{kim2023method} attempted to de-identify real-time videos. They apply de-identification to the landmark area on the object's edge, so it can only de-identify the area that connects the facial feature points, which allows for the preservation of information other than the face. In particular, this method allows for arbitrary selection of de-identification targets.

\textbf{Discussion:}
The main challenge in video de-identification lies in how to ensure seamless modification of the original stream without causing many visual artifacts such as flickering or distortion while preserving temporal consistency across frames. Different methods address it differently, smoothing by spline interpolation over neighboring frames \cite{ciagan} or learned blending mask \cite{gafni2019live}, enforcing resemblance of temporal dynamics to real sequences and adding past frames into input tensors \cite{balaji2021temporally}. Also, data augmentation on input images proves to enhance semantic mapping and robustness on color, lightning \cite{gafni2019live,deepfake}. Besides, further exploiting data redundancy and continuity in videos \cite{wen2022identitymask, savic2023identification, mukherjee2024rid} could provide an interesting line of research.


\section{Applications}
\label{sec:applications}
We present some representative examples of face de-identification applications, which can be divided into targeting specific application scenarios and designing for other computer vision tasks.
\subsection{Specific Areas}
\label{sec:specific-areas}
Zhu et al. \cite{zhu2020deepfakes} proposed to protect patient privacy in medical research by deepfakes. Traditional methods wiped out facial information or body keypoints entirely, which are important in medical diagnoses. Different from other algorithms, there is no need for de-identification in medical research to consider the appearance similarity preservation with the original, so face swapping technology could be applied to eliminate the ethical restrictions of medical data. Similarly, Tian et al. \cite{tian2023generative} generate a context image database with more than 300 fake identity images by StyleGAN and the target is selected by race, age and gender. Then, the target image with a fake identity will be fused with the original image to modify the identity in dental patient images.

Wearable cameras have been widely used for capturing lifevlogs, and a face de-identification system designed for wearable camera images \cite{Puangthamawathanakun2023TowardsFD} has been proposed to replace the original faces with synthetic images. The personal information contained in the face can be deleted, but the face is allowed to remain intact, which health apps can benefit. The proposed system consists of six components: face detection, face alignment, face recognition (ArcFace \cite{arcface}), synthetic face generator, and face swapping (SimSwap \cite{SimSwap2020Chen}). The surrogate face is chosen by calculating an embedding centroid and the cosine similarity should be lower than the threshold to ensure it has a different identity.

FPGAN \cite{lin2021fpgan} could be applied for de-identification on social robot platforms to ensure the data safety of robot users. First, both the workstation and MAT \cite{8750887} (an experimental platform for research on social robots are installed in the FPGAN, and the social robot periodically updated weights with face de-identification on the workstation. When the camera captures face images, it should use the networks to ensure that the user’s visual privacy is not breached.

Lee et al. \cite{lee2021development} proposed a privacy-preserving UAV system to protect identity privacy in the first-person videos taken by UAVs, which just modified face regions instead of blurring or masking the whole area. The trained face-anonymizer could be mounted on the UAV system, so that the people in the video could not be recognized as anyone in the dataset. Additionally, the perception performance required for performing UAV’s essential functions, such as simultaneous localization and mapping, would not be degraded when using such anonymized videos.

``My Face My Choice" (MFMC) argues that access to real identities on social networks should be designed by ``face", with people in different places on the friendship graph seeing different results for the same original image. It is the first complete system designed for using deepfakes to solve the problem of face ownership on social media platforms. When uploading images from the client, users who are connected to the uploader will be selectively tagged, and other faces in the image will be replaced with deepfakes. Tagged friends will get their own unique results based on the friendship, especially outsiders will get the result with all fake faces.

\subsection{Identity-Agnostic Computer Vision Tasks}
\label{sec:advanced-utility}
Considering that it is extremely difficult to retain the utility of all aspects at the same time, some algorithms were designed for specific identity-independent tasks. In this section, we discuss how face de-identification is designed for three broad computer vision tasks: attribute classification, expression analysis, and action detection. Similarly, de-identification can be applied to a wider range by adding corresponding modules to the model design.

\subsubsection{Attribute Classification}
In order to enhance the utility in classification tasks, some de-identification algorithms pay attention to how to achieve identity protection without affecting attributes. A natural thought is to retain deep facial attribute information by loss functions. 
For example, SF-GAN \cite{li2021sf} designed the attribute similarity loss to ensure the generated image can share as much attribute similarity as possible with the original one. A task-agnostic anonymization framework \cite{Barattin2023AttributePreservingFD} was proposed to preserve facial attributes while obfuscating identity by applying identity loss and feature-matching loss \cite{Zheng2021GeneralFR} to optimize latent codes.

The PRN (Privacy Removal Network) \cite{Enhance2023Liu} proposed to remove identity-related information and keep facial attributes with two encoders of different architectures, an internal autoencoder and an external autoencoder. FSN (Feature Selection Network) and PEN (Privacy Evaluation Network) can also be chosen according to the application scenarios. This framework can be used to retain a selected single attribute or multiple attributes and achieve de-identification at the same time, and has higher data availability compared to the $k$-same algorithm.

MAP-DeID \cite{Cao2024MultichannelAP} consists of a multi-channel attribute preservation module and a reconstruction-guided face de-identification module. The attribute preservation module includes three channels, non-facial area, keypoints and facial appearance. The de-identification module implements the modification of the identity while maintaining the authenticity of facial attributes and the naturalness of the image.

Some $k$-anonymity based algorithms implement attribute retention by adding constraints to the selected $k$ images. APFD \cite{attribute-preserved} proposed to estimate the best weights to combine $k$ images by modeling attributes preservation, where the $k$ images selected to share the most similar attributes with the test image to save the corresponding shape and appearance parameters.
The attributes preserving face de-identification method \cite{attribute-preserving} designed to remove identity information while retaining all facial attributes, which can be summarized as: (1) first randomly select $k$ faces which may contain the same or different face attributes with the test image; (2) edit attributes of $k$ selected faces by ELEGANT \cite{elegant} to make them the same as those of the test image; (3) generate the averaged $k$ faces as the de-identified result. GARP-Face \cite{du2014garp} balanced utility preservation in de-identification by determining \textit{Gender, Age and Race} attributes and seeking corresponding representatives to preserve them.

\subsubsection{Expression Recognition}
Some other de-identification methods emphasize obfuscating identity information while maintaining and recognizing facial expressions or emotion, so that the de-identified results can still be used in corresponding classification tasks.

AnonyGAN \cite{dall2022graph} is a landmark-guided de-identification solution that maintains the facial pose of the source face. They achieve facial pose preservation by exploiting a Graph Convolution Network to learn the spatial relations between the source and condition landmarks, which are modeled as a bipartite graph. 
Learned geometrical reasoning is subsequently aggregated to the target's appearance representation. Chen et al. \cite{chen2018vgan} proposed a network that learns an expression-discriminative representation of a face image, disentangled from its identity information. Following a GAN paradigm, the discriminator is designed as a multi-task classifier consisting of three separate auxiliary networks. In addition to distinguishing real face images from synthesized ones, the discriminator is also explicitly imposed to (1) align the expression representation of output face with that of the input face, (2) align the output faces's identity code with the target identity code, earlier fed directly into the generator. Though not specifically designed for face anonymisation, it could be effortlessly adapted to cope with de-identification tasks with minor modifications on the identity discriminator. With a slightly different goal but still under the same approach of amending the discriminator, Aggarwal et al. \cite{aggarwal2020epd} also attached an auxiliary network serving as an emotion verificator to the discriminator, in order to guarantee maximum emotion similarity after de-identification. GANonymization \cite{Hellmann2023GANonymizationAG} achieved keeping facial expressions while anonymizing faces. The pipeline consists of four steps: face extraction, face segmentation, face landmarks detection and re-synthesis, where the face generation network was based on pix2pix architecture to convert facial landmarks to a random and high-quality face image with the same landmarks.


Given an input face, Agarwal et al. \cite{agarwal2021privacy} proposed to select from a set of StyleGAN generated artificial — thus anonymous — faces a proxy face with the ``same" emotion and closest facial pose representation. 
Next, non-biometric attributes are extracted from the input and merged with the chosen proxy face. Intuitively, by directly starting from a proxy face with the same emotion, emotion information is sure to be preserved despite its high computational cost. Similarly, Leibl et al. \cite{De2023Leibl} proposed to generate the synthetic face images at five yaw angles by StyleGAN2 \cite{karras2021alias} to form a library. Secondly, the source face with a similar pose is selected by head pose estimation \cite{Ruiz2017FineGrainedHP}. Finally, the de-identified result is generated by swapping the source face using FSGANv2 \cite{Nirkin2023FSGAN} which can retain pose or expressions. The framework \cite{Shopon2023I} proposed to classify the gender by DeepFace \cite{DeepFace} to select a proxy image from a synthetic image dataset generated by StyleGAN \cite{karras2019style}, and the de-identified result is transformed by SimSwap \cite{SimSwap2020Chen}.

\subsubsection{Action Detection}
Ren et al. \cite{ren2018learning} proposed a GAN-based video face anonymizer that performs de-identification of each person’s face, while maximally maintaining the information used for action recognition. An adversarial paradigm between a face modifier aiming to minimize face identification accuracy, and a face identifier as its discriminator enables anonymization. Action preservation, on the other hand, is attained through an action detector attached to the face modifier, trained on modified frames with ground-truth action labels.

\subsubsection{Face Detection}
Accurate identification of road users, such as pedestrians and cyclists, is important in intelligent transportation systems. Since data regulations require that the datasets used in training must meet certain privacy standards, LFDA \cite{Klemp2023LDFALD} proposes a two-stage approach based on the latent diffusion models. Experimental results prove that LFDA can achieve higher face detection mAP values than other anonymization methods when the face detection model is trained on the de-identified datasets and tested on other non-anonymized datasets.

\section{Evaluation Metrics}
\label{sec:evaluation-metrics}
At present, there are not yet universally acknowledged evaluation criteria for face de-identification. A mere quantitative evaluation framework for de-identification was proposed in \cite{bursic2021quantitative}, from three perspectives of de-identification, expression preservation and photo-reality. On this basis, we sort out the evaluation indicators used in existing papers and present the definition or explanation of the metrics in this section.


\subsection{Privacy Protection}
\begin{itemize}
    \item \textit{Identity Distance (ID)}: Almost all face verification models determine whether two images share the same identity by comparing the distance between their identity embeddings. Therefore, we use the distance between identity-vectors $e_{id}$ extracted from the face recognition model, which can be formulated as,
    \begin{equation}
        \textit{\text{Id-dis}} = Dis(e_{id}(X),e_{id}(\mathcal{F}(X))),
    \end{equation}
    where $X$ indicates the original image, $\mathcal{F}(X)$ represents the de-identification result and the specific form of $Dis$ is determined by the face recognition model used to obtain identity embedding $e_{id}$.
    
    \item \textit{Successful De-identification Rate (SDR)}: In face verification,  if the identity distance exceeds the model's reference threshold, the two images are considered to have different identities. A successful de-identification occurs when the de-identified image’s identity differs from the original. Thus, we compare the identity distance with the threshold to calculate the success rate of de-identification, formulated as:
    \begin{equation}
        \textit{\text{SDR}} = 1-\frac{1}{N}\sum_{i=1}^N f_{ver}(X,\mathcal{F}(X)),
    \end{equation}
    where $f_{ver}=0$ when \textit{Id-dis$>\tau$}, otherwise $f_{ver}=1$, $N$ is the number of testing.
\end{itemize}

\subsection{Utility Preservation}
\subsubsection{General Utility}
\begin{itemize}
    \item \textit{Face detectability (FD)}: 
    The key factor for data utility is that the generated images should appear natural and realistic. This can be measured by assessing how well de-identified faces are detected by face detectors. Two approaches can be used: (1) Face detection rate (\%), which is the proportion of de-identified faces detected by a face detector, or (2) face detectors like \cite{7553523} that provide a confidence score for face detection. Face detectability can then be evaluated by comparing detection scores or softmax probabilities from a face recognition network for both raw and de-identified faces.
    
    \item \textit{Landmark Distance (LD)}: Accurate localization of face regions and keypoints is essential in face modification tasks. It is important that pixel-level landmark deviations remain unaffected during the face de-identification process.

    \item \textit{Peak signal-to-noise ratio (PSNR)}:
    \begin{equation}
        \textit{\text{PSNR}} = 10 \log_{10} (\frac{MAX^2}{MSE}),
    \end{equation}
    where $MAX$ is the maximum pixel value of the image and MSE refers to the mean squared error between two images $I$ and $K$ with the resolution of $m \times n$, calculated by $MSE=\frac{1}{m n} \sum_{i=0}^{m-1} \sum_{j=0}^{n-1}[I(i, j)-K(i, j)]^{2}$.
    
    \item \textit{Structural similarity (SSIM)}:
    We aim to preserve the similarity between the original and de-identified images, rather than randomly altering the face. Specifically, we want to remove only the privacy-related characteristics, but keep the visual similarity, i.e.,  contours and luminous condition. The structural similarity (SSIM) is defined as follows:
    \begin{equation}
        \textit{\text{SSIM}} = l(X,D(X))^\alpha \cdot c(X,D(X))^\beta \cdot s(X,D(X)))^\sigma,
    \end{equation}
    where $l(X,D(X))$, $c(X,D(X))$, and $s(X,D(X))$ denote respectively brightness similarity, contrast similarity and structural similarity, and $\alpha$, $\beta$, $\gamma$ their weighting coefficients, usually set to 1. It is important to note that both PSNR and SSIM just focus on objective image quality evaluation and fail to capture many nuances of human perception \cite{zhang2018unreasonable}.
    
    
    \item \textit{Fréchet Inception Distance (FID)} \cite{heusel2017gans}:
    FID measures the distance between the distribution of real and synthesized images, comparing the visual quality of generated samples to real ones. A lower FID indicates greater similarity between real and generated images. Wang et al. \cite{wang2018video} proposed a variant of FID for video evaluation, which accesses both visual quality and temporal consistency. They modified a video recognition network by removing its last layers, using the remaining part as the ``inception" network. Spatio-temporal feature maps are extracted from each video, and means and covariance matrices are computed for real and synthesized video feature vectors.
    
    \item \textit{Learned Perceptual Image Patch Similarity (LPIPS)} \cite{lpips}: The LPIPS Distance, also known as \textit{perceptual loss}, measures the similarity between two images based on deep features. It was designed to prioritize perceptual similarity, learning the inverse mapping from generated images to ground truth. LPIPS has been shown to correlate better with human perception compared to traditional metrics like MSE, PSNR, SSIM, and FSIM. A lower LPIPS value indicates higher similarity between images, while higher values indicate greater differences.
    \end{itemize}
\subsubsection{Customized Utility}
    \begin{itemize}
    \item \textit{Attribute preservation}:
    To evaluate the proposed model's performance in attribute preservation, we use separate classifiers for each attribute to calculate the accuracy rate (\%) on demographic features such as hair, gender, age, and skin tone.
    \item \textit{Expression preservation}:
    According to the Facial Action Coding System (FACS) \cite{tian2001recognizing}, Action Units (AUs) represent fundamental muscle movements and serve as a proxy for overall facial expressions. Bursic et al. \cite{bursic2021quantitative} proposed extracting AUs from both original and de-identified faces and calculating the root-mean-square error (RMSE) averaged across all AUs. For video de-identification tasks, Pearson's Correlation Coefficient (PCC) is also computed to assess how well AUs correlate over time.

    \item \textit{Temporal Consistency}:
    For video de-identification, preserving temporal consistency is crucial for data utility. The adapted FID metric can measure temporal consistency in synthesized videos, while empirical issues like flickering or identity warping indicate its absence. To quantify temporal coherence, Balaji et al. \cite{balaji2021temporally} proposed the \textit{Identity Invariance Score}, which calculates the identity distance between consecutive frames, averaged over the entire video. This metric assumes that temporal coherence correlates with the invariance of altered identity. Notably, this method can be generalized to assess temporal consistency in other attributes as well.
\end{itemize}

\section{Experimential Comparison}
\label{sec:experiments}
\begin{figure*}[tb]
    \centering
    \includegraphics[width=0.67\linewidth]{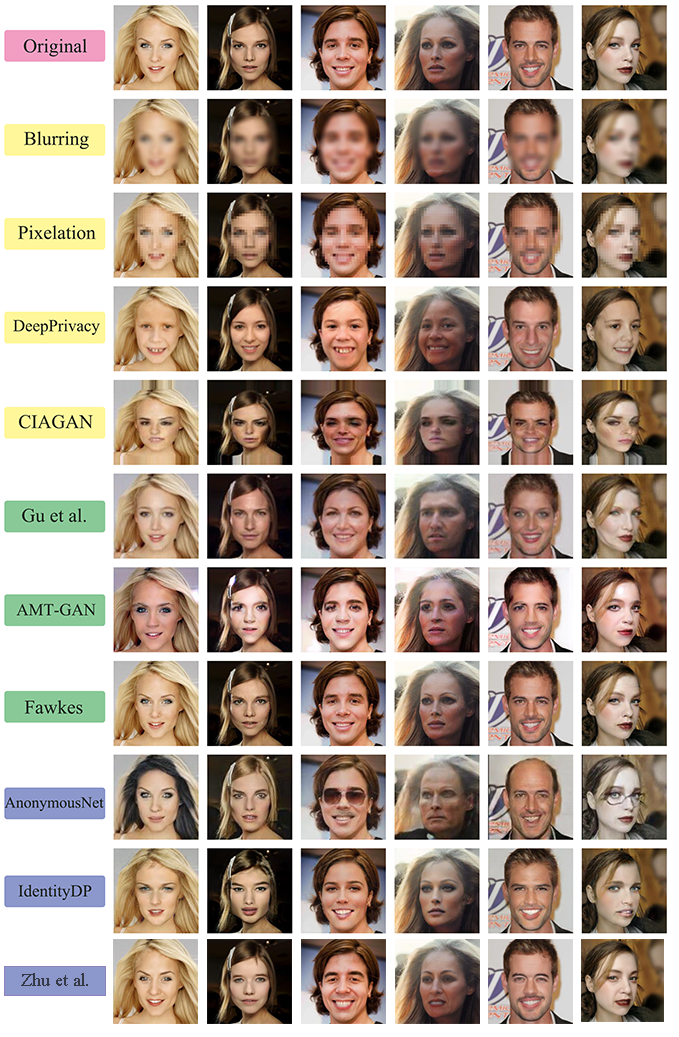}
    \caption{Qualitative comparison results of representative face de-identification methods on CelebA-HQ dataset \cite{karras2018progressive}.}
    \label{fig:qualitative-comparison}
\end{figure*}

We select a few representative algorithms from different categories for qualitative and quantitative experimental comparison, where the selected algorithms are shown in Table~\ref{tab:select-methods}. There is currently no unified dataset for face de-identification, and we choose the CelebA-HQ dataset \cite{karras2018progressive} based on the suitability of the selected methods, which is a new dataset of 256 $\times$ 256 derived from CelebA \cite{liu2015faceattributes} containing 30k images of celebrity faces, and the faces have been cropped based on facial landmarks, so that each of them has a normalized position and rotation. De-identification results generated by different methods are presented in Fig.~\ref{fig:qualitative-comparison}. 

The quantitative comparison results of the privacy and utility indicators in Section~\ref{sec:evaluation-metrics} are shown in Table~\ref{tab:privacy-compare} and Table~\ref{tab:utility-compare}, respectively. Since different pre-processing is required between different algorithms, the ``original image" used in the evaluation is the result after pre-processing, such as image resolution adjustment, for a fairer performance comparison. In the privacy evaluation, we use FaceNet \footnote{\url{https://github.com/davidsandberg/facenet}} and face recognition library \footnote{\url{https://github.com/ageitgey/face_recognition}} to calculate the identity distance. In the utility evaluation, we use dlib library \footnote{\url{https://github.com/davisking/dlib}} for face and landmark detection.

\begin{table*}
  \caption{Selected Methods from Different Categories for Comparison.}
  \label{tab:select-methods}
  \begin{center}
  \begin{small}
  \resizebox{\textwidth}{!}{
  \begin{tabular}{ccccc}
    \toprule
    Process Level & Methods & Category & Dataset & Resolution \\
    \midrule
    \multirow{3}{*}{Pix-Level} & Blurring / Pixelation &  \ref{sec:traditional-methods} & - & -  \\
    & DeepPrivacy \cite{deepprivacy} & \ref{sec:GAN-based-face-generation} & WIDER-Face  & 128 $\times$ 128\\
    & CIAGAN \cite{ciagan} & \ref{sec:GAN-based-face-generation} & CelebA & 128 $\times$ 128\\
    \hline
    \multirow{3}{*}{Rep-Level} 
    & Gu et al. \cite{gu2020password} & \ref{modifiedGANs} & CASIA / LFW / FFHQ & 128 $\times$ 128\\
    & AMT-GAN \cite{hu2022protecting} & \ref{sec:adversarial-pertubation} &  Makeup Transfer / CelebA-HQ / LAND & 256 $\times$ 256\\
    & Fawkes \cite{shan2020fawkes} & \ref{sec:adversarial-pertubation} & WebFace / VGGFace2 & 224 $\times$ 224\\
    \hline
    \multirow{3}{*}{Sem-Level} & AnonymousNet \cite{anonymousnet} & \ref{sec:attributes} & CelebA  &  128 $\times$ 128\\     
    & IdentityDP \cite{wen2022identitydp} &  \ref{sec:identity-representation} & CelebA-HQ & 256 $\times$ 256\\
    & Zhu et al. \cite{Zhu2024IdentityConsistentVD} & \ref{sec:identity-representation} & VoxCeleb & 256 $\times$ 256 \\
  \bottomrule
\end{tabular}}
\end{small}
\end{center}
\end{table*}

\begin{table*}[tb]
\caption{Comparison of the selected representative algorithms on CelebA-HQ \cite{karras2018progressive} under privacy protection effectiveness metrics, where the \textcolor{R1}{red} one represents the best and the \textcolor{R2}{blue} one indicates the second.}
\begin{center}
\resizebox{\textwidth}{!}{
\begin{tabular}{c c | c c| c c |c c}
\toprule
\multirow{2}{*}{Process-Level} & \multirow{2}{*}{Method} & \multicolumn{2}{c|}{FaceNet (VGGFace2)} & \multicolumn{2}{c|}{FaceNet (CASIA)} & \multicolumn{2}{c}{Face Recognition}\\
 & & ID$\uparrow$ & SDR$\uparrow$ & ID$\uparrow$ & SDR$\uparrow$ & ID$\uparrow$ & SDR$\uparrow$ \\
\hline
\multirow{4}{*}{Pix-Level} 
& Blurring ($r$=20) & 1.082 & 0.785 & 1.051 & 0.710 &0.510 & 0.371\\
& Pixelation (4$\times$4) & 0.938 & 0.348  & 0.940 & 0.362 &0.519 &0.171 \\
& DeepPrivacy \cite{deepprivacy} & 1.206 & \textcolor{R1}{0.940} & 1.172 & 0.897 &0.749 & 0.948  \\
& CIAGAN \cite{ciagan} & 0.591 & 0.412 & 0.562 & 0.352 & 0.353 & 0.452 \\
\hline
\multirow{3}{*}{Rep-Level} 
& Gu et al. \cite{gu2020password} & \textcolor{R1}{1.261} & \textcolor{R2}{0.936} & \textcolor{R1}{1.238} & \textcolor{R1}{0.935} & \textcolor{R1}{0.838}  & \textcolor{R1}{0.978} \\
& AMT-GAN \cite{hu2022protecting} & 0.939 & 0.322 & 0.915 & 0.286 & 0.556 & 0.405\\
& Fawkes (low-level) \cite{shan2020fawkes} & 0.468 & 0.0 & 0.495 & 0.0  & 0.318 & 0.0\\
\hline
\multirow{3}{*}{Sem-Level} & AnonymousNet \cite{anonymousnet} & 0.724 & 0.452 & 0.806 & 0.395 & 0.489 & 0.534\\
& IdentityDP ($\varepsilon$=0.57) \cite{wen2022identitydp} & \textcolor{R2}{1.225} & 0.912 & \textcolor{R2}{1.184} & \textcolor{R2}{0.906} & 0.753 & \textcolor{R2}{0.965}\\
& Zhu et al. (T=100) \cite{Zhu2024IdentityConsistentVD}  & 1.214 & 0.903 & 1.152 & 0.873 & \textcolor{R2}{0.756} & 0.949\\
\bottomrule
\end{tabular}}
\label{tab:privacy-compare}
\end{center}
\end{table*}

\begin{table*}[tb]
\caption{Comparison of the selected representative algorithms on CelebA-HQ \cite{karras2018progressive} under utility preservation metrics, where the \textcolor{R1}{red} one represents the best and the \textcolor{R2}{blue} one indicates the second.}
\begin{center}
\resizebox{\textwidth}{!}{
\begin{tabular}{c c c c c c c c c}
\toprule
Process-Level & Method  &FD$\uparrow$ &  LD$\downarrow$ & PSNR$\uparrow$ & SSIM$\uparrow$ & FID$\downarrow$ & LPIPS$\downarrow$ & MAE$\downarrow$ \\
\midrule
\multirow{4}{*}{Pix-Level} 
&Blurring ($r$=20) & 0.869 & 1.816 & 25.402 & 0.792 & 74.920 & 0.246 & 0.623\\
&Pixelation (4$\times$4) & 0.971 & \textcolor{R2}{1.398} & \textcolor{R2}{25.927} & 0.777 & 42.747 &0.276 & 0.657 \\
&DeepPrivacy \cite{deepprivacy} & \textcolor{R1}{1.0} & 2.225 & 22.084 & 0.823 & 23.412 &0.383 & 0.582 \\
&CIAGAN \cite{ciagan} & 0.979 & 4.355 & 14.342 & 0.404 & 30.703  & 0.314 & 1.059 \\
\hline
\multirow{3}{*}{Rep-Level} 
&Gu et al. \cite{gu2020password} & \textcolor{R2}{0.998} & 3.608 & 23.248 & 0.744 & 42.465 & 0.259 & 0.699 \\
&AMT-GAN \cite{hu2022protecting} &0.995 & 2.305 & 20.826 & 0.773 &35.390 & \textcolor{R2}{0.105} & 0.816\\
&Fawkes (low-level) \cite{shan2020fawkes} & 0.990 & \textcolor{R1}{0.426} &\textcolor{R1}{43.960} &\textcolor{R1}{0.994} &\textcolor{R1}{1.092} &\textcolor{R1}{0.005} & \textcolor{R1}{0.269} \\
\hline
\multirow{3}{*}{Sem-Level} & AnonymousNet \cite{anonymousnet} & 0.867 & 3.415 & 20.302 & 0.798 & 55.047 & 0.656 & 0.745\\
& IdentityDP ($\varepsilon$=0.57) \cite{wen2022identitydp} & 0.997 & 1.856 & 24.013 & 0.861 & \textcolor{R2}{23.217} & 0.231 & 0.416\\
& Zhu et al. (T=100)\cite{Zhu2024IdentityConsistentVD}  & 0.992 & 2.185 &25.142 &\textcolor{R2}{0.878} & 30.495 & 0.311 &\textcolor{R2}{0.402} \\
\bottomrule
\end{tabular}}
\label{tab:utility-compare}
\end{center}
\end{table*}

On the basis of the experimental results, we summarize the advantages and disadvantages of each technology category as follows.
\begin{itemize}
    \item The traditional methods are simple to operate, but the protection performance is limited. When the added disturbance degree is small, it has little impact on image quality but poor protection performance. It is found that increasing the disturbance degree contributes higher protection effectiveness, but the image quality and utility will be damaged greatly, such as the inability to detect faces. 
    
    \item Deep learning-based methods can deal with the trade-off better with higher-quality de-identification results. \textit{Generative Models-Based Face Generation} methods replace the whole face region at pixel level, they can't consider the similarity with the original. \textit{Generative Models-Based Face Editing} methods at the representation level achieve de-identification by designing loss functions, they often cannot adjust the protection degree after model training, which has insufficient flexibility. 
    
    \item For the approaches of semantic-level, \textit{attribute-based} methods have diminishing effectiveness in privacy because there is no clear correspondence between attributes and identity. Generally speaking, \textit{identity-based} algorithms have the best performance in controlling the trade-off, and achieves more effective protection through minor but more targeted modification.
    
    \item Although adding adversarial noise seems like an ideal protection method, which can affect the recognition accuracy without affecting the image content. However, it is hard to apply to various face recognition models in practice. For example, although Fwakes \cite{shan2020fawkes} performs well for Amazon Rekognition Face Verification API while it fails to obtain successful de-identification results in our experiments. On the other hand, the designed noise cannot achieve an ``invisible" quality in the current state of the art. There still exists perceptual noise texture even if being combined with other face painting tasks like AMT-GAN \cite{hu2022protecting}.
    
\end{itemize}

\section{Future Directions}
\label{sec:future-direction}
In this section, we provide a summary of the main challenges and open issues associated with face de-identification methods and then point out future directions in our view. 

\textbf{Targeted Modification.}
Based on the summary and experimental comparison, we think that more fine-grained processing on the part targeted at identity can achieve both more effective protection effects and improved utility. Obviously, if the identity-related parts can be separated and operated, better protection can be achieved under the same disturbance degree. How to disentangle identity and attribute is still an active area with much ongoing research work, including exploring a more explicit latent space, more comprehensive identity representation \cite{wang2021hififace}, the introduction of contrast loss \cite{jeong2021ficgan} in training, etc.

\textbf{Controllable and Reversible Privacy.} Different scenarios require varying levels of de-identification, and user needs can differ even within the same context. Therefore, an ideal face de-identification method should be adjustable and controllable to suit diverse applications. One way to increase flexibility is to separate the protection process from network training. In certain cases, such as crime tracking, the original images may be preferable to the de-identified ones. Currently, most research focuses on protection, with few methods offering recoverable de-identification. Exploring restoration processes in the future could be highly valuable.

\textbf{Generalization Across Diverse Applications.} Some algorithms rely on the pre-trained face recognition model to obtain identity embeddings, especially \textit{Generative Models-Based Identity Modification} algorithms and adversarial perturbation, where the former group has better generalization and the latter is often effective only for specific face recognition models. In addition, deep neural networks are always trained in a data-driven manner, so the performance will be influenced greatly by datasets. There is often no uniform standard for face images in practical applications, so it is very challenging to break through the dataset deviation and produce satisfactory results for different inputs.

\textbf{Privacy Guarantees.} Provable privacy is another important aspect of face de-identification methods. Existing privacy metrics like $k$–anonymity is based on strong assumptions and fail to handle real-world problems. It is worth considering how to better combine privacy theory with deep learning technology to enhance de-identification algorithms' measurement and interpretability. Additionally, due to the particularity of image data, some improvement based on traditional privacy mechanisms is required to be adapted to the specific tasks.

\textbf{Evaluation Standards.} Although we summarize the evaluation metrics commonly used in current de-identification algorithms from both privacy and utility in Section~\ref{sec:evaluation-metrics}, they still cannot comprehensively quantify the algorithm performance. Most of the utility metrics are based on the similarity between the de-identified results and the original image. To the best of our knowledge, there is no generally accepted evaluation system or indicator for de-identification. Proposing new evaluation criteria for the characteristics of de-identification algorithms is a driving force for future development in this field.

\textbf{Systematic Design for Scalability and Real-time Performance.} There is now a growing need for scalable and efficient de-identification methods that can handle high-resolution images and video streams in real-time. Future work should prioritize the development of lightweight algorithms that maintain robust privacy protection without sacrificing computational efficiency. Techniques such as model compression, pruning, and edge computing could be leveraged to enhance scalability and deployment across resource-constrained environments.

\textbf{Broader Multi-modal Privacy Protection Systems.} The increasing sophistication of AI systems means that facial data is often integrated with other biometric identifiers, such as voice, gait, or even behavioral patterns. Future research should investigate how face de-identification techniques can be integrated with multi-modal privacy protection frameworks that secure not only visual data but also other sensitive information. For instance, combining face de-identification with voice anonymization or gesture obfuscation could create a more comprehensive and effective privacy solution, particularly in scenarios where multiple forms of biometric data are simultaneously captured.

\section{Conclusion}
\label{sec:conclusion}
This survey provided a comprehensive survey of face de-identification for protecting sensitive facial information, especially identity. Based on the proposed taxonomy of image-processing level, we have categorized existing methods in \textit{pixel-level}, \textit{representation-level} and \textit{semantic-level}. With the increasing importance of privacy preservation in the era of AI-driven facial recognition systems, de-identification methods have become essential for protecting individuals' sensitive data while maintaining the usability of images for non-identification tasks. We reviewed state-of-the-art methods and proposed an evaluation framework that measures the effectiveness of these methods in terms of privacy protection and image utility.

In addition, we present practical applications that include specific domains and identity-independent computer vision tasks. Furthermore,  we conduct qualitative and quantitative experimental comparisons for each category of algorithms to better demonstrate their performance. Our experimental comparison of representative methods reveals that deep learning-based approaches, especially those leveraging Generative Adversarial Networks (GANs) and diffusion models, have significantly improved the trade-off between privacy protection and utility preservation. However, challenges remain in achieving high fidelity while ensuring robust privacy, particularly under adversarial conditions or complex image distortions. For example, while semantic-level methods offer more flexibility by manipulating high-level features, they often struggle with consistency across varied scenarios. In contrast, pixel-level methods, though more straightforward, frequently degrade image quality, limiting their practical applications.

Looking ahead, several research challenges need to be addressed. First, more robust methods are needed to provide fine-grained, controllable, and guaranteed privacy with image utility across diverse contexts. Second, the field would benefit from a deeper exploration of system designs for more scalable and real-time applications. Third, the integration of privacy-preserving techniques with multi-modal privacy protection systems could create a more comprehensive and effective privacy solution.

In conclusion, while significant progress has been made, face de-identification remains a vibrant area of research with many open questions. This survey offers a foundation for further exploration and highlights both the opportunities and the obstacles in designing next-generation de-identification systems that are robust, efficient, and secure.



\bibliographystyle{IEEEtran}
\bibliography{references}

\newpage

\end{document}